\documentclass[pmlr]{jmlr}
\RequirePackage{graphicx}
 \usepackage{booktabs}
\usepackage{longtable}
\usepackage{multirow} 
\usepackage{algorithm}
\usepackage{algorithmic}
\makeatletter
\def\set@curr@file#1{\def\@curr@file{#1}} 
\makeatother
\usepackage[load-configurations=version-1]{siunitx} 

\theorembodyfont{\upshape}
\theoremheaderfont{\scshape}
\theorempostheader{:}
\theoremsep{\newline}

\jmlrvolume{298}
\jmlryear{2025}
\jmlrworkshop{Machine Learning for Healthcare}

\title[ECG-Byte]{ECG-Byte: A Tokenizer for End-to-End Generative Electrocardiogram Language Modeling}

\author{%
\Name{William Jongwon Han} \Email{wjhan@andrew.cmu.edu}\\
\addr Carnegie Mellon University, USA
\AND
\Name{Chaojing Duan} \Email{chaojing.duan@ahn.org}\\
\addr Allegheny Health Network, USA
\AND
\Name{Michael A. Rosenberg} \Email{michael.a.rosenberg@cuanschutz.edu}\\
\addr University of Colorado, USA
\AND
\Name{Emerson Liu} \Email{emerson.liu@ahn.org}\\
\addr Allegheny Health Network, USA
\AND
\Name{Ding Zhao} \Email{dingzhao@andrew.cmu.edu}\\
\addr Carnegie Mellon University, USA
}

\begin{document}

\maketitle

\begin{abstract}
Large Language Models (LLMs) have demonstrated exceptional versatility across domains, including applications to electrocardiograms (ECGs). A growing body of work focuses on generating text from multi-channeled ECG signals and corresponding textual prompts. Existing approaches often involve a two-stage process: pretraining an ECG-specific encoder with a self-supervised learning (SSL) objective, followed by finetuning an LLM for natural language generation (NLG) using encoder-derived features. However, these methods face two key limitations: inefficiency due to multi-stage training and challenges in interpreting encoder-generated features.
To overcome these issues, we propose \textbf{ECG-Byte}, an adapted byte pair encoding (BPE) tokenizer pipeline for autoregressive language modeling of ECGs. \textbf{ECG-Byte} compresses and encodes ECG signals into tokens, enabling direct end-to-end LLM training by combining ECG and text tokens. This approach enhances interpretability, as ECG tokens can be directly mapped back to the original signals. Leveraging \textbf{ECG-Byte}, we achieve competitive NLG performance while training \textbf{3 times faster} and using just \textbf{48\% of the data} required by traditional two-stage methods. All code is available at \href{https://github.com/willxxy/ECG-Byte}{\texttt{github.com/willxxy/ECG-Byte}.}
\end{abstract}

\section{Introduction}
\label{sec:intro}
Cardiovascular diseases (CVDs) are the leading cause of global mortality, with 17.9 million lives taken each year and increasing \citep{worldhealthorganization_2024_cardiovascular}.
Due to their readily available, noninvasive and information dense nature, 12-lead ECGs are first-line diagnostic tools for screening/evaluation of potential CVDs. 
However, accurate ECG analysis is limited in places where ECG expertise is not accessible, exacerbated by the decline and lack of available cardiac electrophysiologists especially in rural areas \citep{johnson_2024_counties}.

\begin{figure*}[!htp]
\centering
\includegraphics[width=1\linewidth]{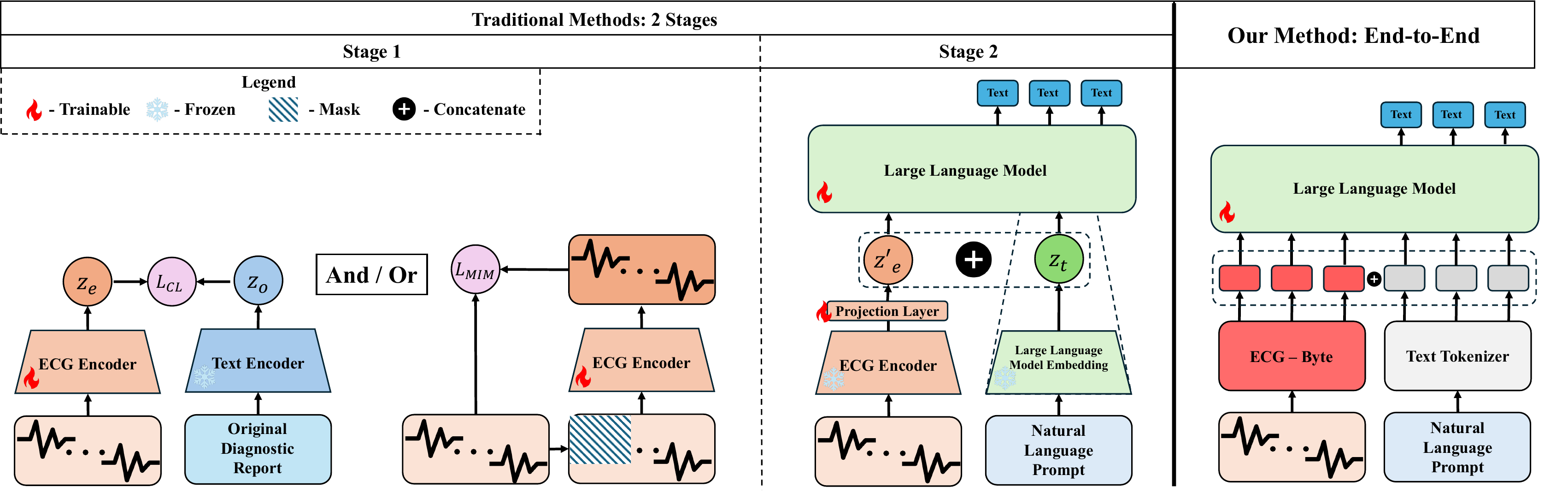}
\caption{Comparison of traditional and our approach for ECG language modeling. Traditional methods follow a two-stage process: (i) training a 12-lead ECG encoder with self-supervised objectives—contrastive learning ($L_{CL}$ between ECG ($z_e$) and the textual diagnostic report ($z_o$)) and/or masked image modeling ($L_{MIM}$)—to learn robust latent features; and (ii) mapping these ECG features ($z'_e$) to a shared representation space via a learnable projection layer, then concatenating them with textual prompt vectors ($z_t$) as input to a large language model (LLM) for generation. In contrast, our method trains an LLM end-to-end using \textbf{ECG-Byte} as an ECG tokenizer.}
\label{Fig:model}
\end{figure*}

The aforementioned facts calls attention to the need for accessible, accurate, and efficient automation of ECG analysis through deep learning.
Deep learning has reached expert level performance in certain tasks for CVD detection using ECGs \citep{rajpurkar2017cardiologistlevelarrhythmiadetectionconvolutional, hannun_2019_cardiologistlevel}.
However, most previous works have succumbed to a crude classification of hard CVD labels \citep{nonaka2020electrocardiogram_ecg_data_aug, Martin2021RealtimeFS, Strodthoff2021DeepLF}.
A problem with this approach is that ECGs often do not exclusively fall into one diagnostic category, instead, there may be many soft labels annotated by expert physicians and the accumulation of these soft labels allow a more detailed, nuanced, and clinically useful interpretation of the ECG \citep{Singstad2022.11.16.22282373, app11167758}.

The recent emergence of Large Language Models (LLMs) enables a generative, flexible approach to ECG analysis. Recent works have adopted this method \citep{tang2024electrocardiogramreportgenerationquestion, zhao2024ecgchatlargeecglanguagemodel}, treating multi-channel ECGs as images by first pretraining an ECG-specific encoder with a self-supervised learning (SSL) objective and then finetuning an LLM for natural language generation (NLG). However, this two-stage process has two key drawbacks: (1) training inefficiencies, as pretraining an effective encoder demands significant computational resources due to large datasets, model sizes, and long training times, and (2) interpretability challenges, since the encoder’s latent feature vectors cannot be mapped back to the original signal.

In this study, we introduce \textbf{ECG-Byte}, a byte pair encoding (BPE) tokenizer \citep{Gage1994ANA} designed for end-to-end training of generative ECG language models. 
Building on prior work that converts continuous values to discrete tokens \citep{chen2022pix2seq, han2024interpretationintracardiacelectrogramstextual}, we use quantization to represent amplitude ranges as discrete symbols and obtain string representations of ECGs. 
We then directly finetune an LLM for autoregressive natural language generation (NLG), conditioning text output on both the text prompt and tokenized signal. 
Our end-to-end approach is competitive for conditional NLG, with \textbf{3 times faster} total training time and \textbf{approximately 48\% of the data} compared to two-stage pretraining methods. 
Moreover, since the tokenized ECG can be directly traced back to the original signal, token-level attention-based visualizations become interpretable.

Our contributions are summarized below:\\
1. We introduce \textbf{ECG-Byte}, a tokenizer that compresses discretized symbolic representations of ECGs using BPE, enabling direct end-to-end training without the need for an ECG-specific encoder.\\
2. We empirically show the efficiency of our method and present competitive performances against conventional two-stage pretraining approaches, proposing a new paradigm for conditional NLG with ECGs.\\
3. We perform an interpretability study on \textbf{ECG-Byte} and the LLM, examining how \textbf{ECG-Byte} merges ECG signals and using attention visualizations to analyze how the LLM processes ECGs and text.

\subsection*{Generalizable Insights about Machine Learning in the Context of Healthcare}
The motivation and generalizable insights of \textbf{ECG-Byte} derive from Vision-Language Models (VLMs) \citep{liu2023visualinstructiontuning, liu2024improvedbaselinesvisualinstruction}, where image encoders are pretrained on large-scale data and later applied for visual understanding tasks. While recent ECG research has focused on developing specialized encoders for classification \citep{na2024guidingmaskedrepresentationlearning, liu2024zeroshotecgclassificationmultimodal, kiyasseh2021clocscontrastivelearningcardiac} and some works like ECG-Chat \citep{zhao2024ecgchatlargeecglanguagemodel} have shown potential for using these encoders in NLG, \textbf{ECG-Byte} demonstrates that a simpler approach can achieve competitive results. Our key insight is that compressing ECG data via a rule-based method (BPE tokenization) eliminates the need for complex encoder training pipelines that often involve additional overhead due to the size of the encoder, multiple loss objectives and carefully guided feature extraction \citep{gopal20213kgcontrastivelearning12lead, oh2022leadagnosticselfsupervisedlearninglocal}. Our analysis reveals that \textbf{ECG-Byte} effectively preserves clinically relevant signal components (P wave, QRS complex, T wave) within its tokenization scheme. Beyond cardiovascular applications, this successful adaptation of BPE for time series data suggests a promising approach for processing other physiological signals such as electroencephalograms (EEGs), photoplethysmograms (PPGs), or electromyograms (EMGs) in generative tasks with LLMs.

\section{Related Works}
\subsection{Deep learning for ECGs}
There has been a plethora of works utilizing deep learning for processing ECGs for classification \citep{rajpurkar2017cardiologistlevelarrhythmiadetectionconvolutional, hannun_2019_cardiologistlevel, nonaka2020electrocardiogram_ecg_data_aug, Martin2021RealtimeFS, Strodthoff2021DeepLF}.
Most of these works utilize either convolutional neural networks (CNNs) \citep{rajpurkar2017cardiologistlevelarrhythmiadetectionconvolutional} or transformers \citep{choi2023ecgbert} and exhibit excellent performance at classification tasks.
There have also been some efforts to frame classification as a retrieval task in order to recover cases similar to the given ECG \citep{tang2024electrocardiogramreportgenerationquestion, pmlr-v225-qiu23a}.
However, the retrieval approach may struggle with rare or unique ECG patterns that lack good matches in the existing database.
Additionally, both classification and retrieval tasks may be crude formulations of processing ECGs, since ECGs typically exhibit many characteristics of overlapping CVDs.

\subsection{Large Language Models for ECGs}
Generative Large Language Models (LLMs) have given the opportunity to take a softer and more clinically similar approach in processing ECGs by generating physician-vetted clinical statements \citep{qiu-etal-2023-transfer, tang2024electrocardiogramreportgenerationquestion, wan2024meitmultimodalelectrocardiograminstruction, cardiogpt, zhao2024ecgchatlargeecglanguagemodel}.
Previous works have largely focused on representing ECG data by feeding the raw signal into a neural network encoder to output a latent representation, which is then used as input for an LLM \citep{zhao2024ecgchatlargeecglanguagemodel, wan2024meitmultimodalelectrocardiograminstruction, tang2024electrocardiogramreportgenerationquestion}. 
Recent studies have explored using printed ECGs as images for natural language generation (NLG) \citep{liu2024teachmultimodalllmscomprehend, Khunte2024.02.17.24302976}. 
However, our study focuses solely on methods that utilize the direct ECG signal.
In order to obtain robust latent representations of the ECG signal, an encoder is first trained on a self-supervised learning (SSL) objective (e.g., contrastive learning, masked language/image modeling). 
Although model performance in terms of label classification has been excellent when using these approaches, we want to be able to generate soft labels akin to clinical notation since ECGs often have overlapping and non-mutually exclusive descriptors. 
While some efforts have been made in this direction \citep{tang2024electrocardiogramreportgenerationquestion, zhao2024ecgchatlargeecglanguagemodel}, they face challenges in efficiency (requiring two stages of training) and interpretability (as the latent feature vectors from the ECG encoder are not interpretable). 
In our work, we challenge these two-stage pretraining approaches by transforming the ECG into tokens using \textbf{ECG-Byte} and directly training an LLM for NLG.

\subsection{Byte Pair Encoding for Domains Outside of Language}
The Byte Pair Encoding (BPE) algorithm was introduced by \citet{Gage1994ANA} for data compression and later adapted for natural language processing (NLP) \citep{sennrich2016neuralmachinetranslationrare}. 
It is favored for tokenization in popular language models \citep{grattafiori2024llama3herdmodels, brown2020languagemodelsfewshotlearners} due to its efficiency and robustness to rare words. 
Beyond language, BPE has been applied to modalities such as molecular graphs \citep{shen2024graphbpemoleculargraphsmeet}, electroencephalograms (EEG) \citep{mykolaklymenko_2023_bytepair}, and other physiological signals \citep{tavabi2021patterndiscoverytimeseries} for classification. 
Most recently, \citet{tahery2024heartbertselfsupervisedecgembedding} used quantization and BPE to compress ECG signals as inputs to a BERT \citep{devlin2019bert} model for self-supervised learning, though only for classification. 
We argue that classification alone may limit ECG interpretation and therefore leverage these representations for generative diagnosis. 
Additionally, previous works employ a pre-existing BPE tokenizer \citep{tahery2024heartbertselfsupervisedecgembedding} based on SentencePiece \citep{kudo2018sentencepiecesimplelanguageindependent} without further analyzing \textit{how} the BPE algorithm merges ECGs.
Inspired by HuggingFace \citep{wolf2020huggingfaces}, we develop \textbf{ECG-Byte}, a custom Rust-based BPE tokenizer for ECGs. We conduct an extensive analysis of merged token usage, the distribution of tokenized ECG lengths, and provide visualizations mapping compressed tokens to the original ECG signal.

\section{Methods}
\label{sec:method}
This section provides detailed information on the datasets, preprocessing, ECG signal encoding with \textbf{ECG-Byte}, and LLM training for NLG.

\subsection{Dataset and Preprocessing}
\paragraph{Dataset} In this study, we use variants of the MIMIC-IV ECG \citep{https://doi.org/10.13026/4nqg-sb35} and PTB-XL datasets \citep{wagner_ptb-xl_2020} for NLG.
We use MIMIC-IV ECG pretraining curated by \citet{zhao2024ecgchatlargeecglanguagemodel} that contains question prompts generated by GPT-4o alongside the ECG and clinical notes.
Additionally, we use the ECG-QA dataset \citep{oh2023ecgqacomprehensivequestionanswering}, a dataset that uses the ChatGPT API to generate naturalistic, clinically relevant question and answer pairs about the ECG signals from the MIMIC-IV ECG and PTB-XL datasets.
The baselines we compare our results with all utilize the \textit{single-verify}, \textit{single-choose}, and \textit{single-query} categorized questions from the ECG-QA dataset. 
The ECG signals collected from both datasets (i.e., MIMIC-IV ECG and PTB-XL) are sampled at 500 Hz for 10 seconds, resulting in a 5000 length, 12 lead ECG.

\paragraph{Preprocessing}
All datasets are preprocessed uniformly. For the MIMIC-IV ECG, we first reorder leads to match the PTB-XL format (from [I, II, III, aVR, aVF, aVL, V1–V6] to [I, II, III, aVL, aVR, aVF, V1–V6]). 
Powerline interference is removed using bidirectional notch filters at 50 Hz and 60 Hz (Q=30). 
A fourth-order Butterworth bandpass filter (0.5–100 Hz) isolates ECG components, while a bidirectional fourth-order highpass filter (cutoff 0.05 Hz) mitigates baseline wander. 
We then apply Daubechies-6 (db6) wavelet denoising at level 4, using a soft threshold based on the median absolute deviation of the detail coefficients. 
The signal is downsampled from 500 Hz to 250 Hz and segmented into non-overlapping 2-second windows for model input—except during tokenizer training, where the full 10-second signal is retained to avoid discontinuity. 
Finally, global 1st and 99th percentiles (from 300,000 samples) are recorded for later normalization in training \textbf{ECG-Byte}.

\subsection{ECG as Bytes}\label{ecg-byte-exp}
\paragraph{Sampling}
Following established practices in NLP \citep{dagan2024gettingtokenizerpretrainingdomain}, we train \textbf{ECG-Byte} on a representative subset of the total dataset, selected using stratified sampling based on morphological clustering.
To extract features from each unsegmented ECG, we compute statistical measures, frequency and time domain features, morphological characteristics, and wavelet coefficients. Principal Component Analysis (PCA) \citep{WOLD198737} is applied for dimensionality reduction, retaining 95\% of the variance, followed by feature scaling.
The optimal number of clusters is determined using the Elbow Method and Silhouette Analysis \citep{ROUSSEEUW198753}, with the smaller result chosen. K-means clustering \citep{macqueen_1967_some} is then applied to the scaled PCA-transformed features. If K-means fails to yield distinct clusters, DBSCAN \citep{10.5555/3001460.3001507} is used as a fallback.
Stratified sampling is performed by randomly selecting ECGs from each cluster in proportion to its size, resulting in a total sample of 200,000 ECGs for training \textbf{ECG-Byte}.

\paragraph{Quantization}
To ensure consistency across ECG signals, we normalize each input by scaling it to a fixed range and encoding it into a symbolic representation. Let \( X \in \mathbb{R}^{C \times T} \) denote an ECG signal matrix, where $C$ is the number of ECG leads and $T$ represents the number of sampled time points per lead. In this study, $C = 12$ and $T = 500$ unless specified otherwise. 
Let \( p_1 \) and \( p_{99} \) represent the 1st and 99th percentiles of \( X \) across all leads and time points sampled earlier during preprocessing, respectively. 
The normalization process is defined as follows:
\begin{equation}
X_{\text{norm}} = \frac{X - (p_1 - \epsilon_1)}{(p_{99} + \epsilon_1) - (p_1 - \epsilon_1) + \epsilon_2}
\end{equation}
where \( \epsilon_1 = 0.5 \) is a constant to make up for the sampled percentiles and \( \epsilon_2 = 10^{-6} \) is a small constant added to prevent division by zero. This transformation shifts and scales \( X \) so that the normalized values fall within the range [0, 1]. We then apply clipping to ensure that values remain strictly within this range.
Inspired by previous works \citep{mykolaklymenko_2023_bytepair, tavabi2021patterndiscoverytimeseries, chen2022pix2seq, han2024interpretationintracardiacelectrogramstextual}, we quantize \( X_{\text{norm}} \) into discrete levels for a symbolic representation. 
Let \( \mathcal{A} \) be the set of 26 symbols, corresponding to the lowercased letters in the English alphabet, $\mathcal{A} = \{ \text{a}, \text{b}, \ldots, \text{z} \}$.
The alphabet size \( |\mathcal{A}| = 26 \) defines the number of discrete levels. 
We scale and floor \( X_{\text{norm}} \) to integer values, then take the minimum between the floored value and the maximum number of bins as the following:
\begin{equation}
X_{\text{quant}} = \text{min}(\lfloor X_{\text{norm}} \times |\mathcal{A}|\rfloor, (|\mathcal{A}| - 1))
\end{equation}
Finally, each integer value in \( X_{\text{quant}} \) is mapped to a corresponding symbol in \( \mathcal{A} \) to yield the symbolic signal, which serves as a discrete representation of the ECG.
After transforming each ECG signal instance into its symbolic form, we first flatten each symbolic ECG instance \( X_{\text{quant}}^{(i)} \) into a 1-dimensional sequence of symbols $X_{\text{symb}}^{(i)}$, where $X_{\text{symb}}^{(i)} \in \mathcal{A}^{CT}$, \( i \) indexes over all instances in the dataset, and \( X_{\text{symb}}^{(i)} \) is the flattened sequence of symbols of length \( C \cdot T \).
Next, we concatenate all flattened instances \( X_{\text{symb}}^{(1)}, X_{\text{symb}}^{(2)}, \ldots, X_{\text{symb}}^{(N)} \) across the entire dataset to form a single, long symbolic sequence $X_{\text{concat}}$, where $X_{\text{concat}} \in \mathcal{A}^{NCT}$, and \( N \) is the total number of instances in the dataset.
The concatenated symbolic sequence \( X_{\text{concat}} \) of length \( N \cdot C \cdot T \) is then used to train \textbf{ECG-Byte}.

\paragraph{ECG-Byte Training Process}
After obtaining the string representation \( \mathbf{X}_{\text{concat}} \) of the ECG dataset, we train \textbf{ECG-Byte} to compress the discretized ECG signals by iteratively merging the most frequent byte pairs into single tokens, following the BPE algorithm \citep{Gage1994ANA}.
The process starts by converting \( \mathbf{X}_{\text{concat}} \) into a vector of token IDs derived from 8-bit byte values (stored as 32-bit unsigned integers) and initializing a vocabulary map (\texttt{vocab}) for string representations of bytes and a \texttt{vocab\_tokens} map to encode bytes as singleton lists. 
IDs and \texttt{vocab} are initialized to cover the full byte range (0--255), mapping symbols in \( \mathcal{A} \) to ASCII values (97--122), while reserving other byte values for unknown bytes. 
As merging proceeds, new tokens are assigned unique integer IDs starting from 256, acting as abstract labels for progressively larger token units.
For each merge iteration, \textbf{ECG-Byte} calculates adjacent byte pair frequencies using a parallelized \texttt{get\_stats} function, efficiently aggregating counts via a fold-and-reduce strategy. 
The most frequent pair is identified as the ``best pair" to merge, and the \texttt{merge} function replaces occurrences of this pair in the ID vector with a new token ID, extending the vocabulary and updating \texttt{vocab\_tokens} accordingly. 
This process repeats until the specified number of merges is reached or no pairs remain.
The output includes the encoded ID vector, the extended vocabulary map, and a history of merge operations. 
Existing tokenizers, such as SentencePiece \citep{kudo2018sentencepiecesimplelanguageindependent} or HuggingFace \citep{wolf2020huggingfaces}, were not used due to their complexity and integration issues, which hindered interpretability. 
\textbf{ECG-Byte}, implemented in Rust for speed, provides a lightweight, flexible framework for representing ECG signals as discrete tokens while drawing inspiration from HuggingFace's tokenizer \citep{wolf2020huggingfaces}. 
We provide the detailed pseudocode of the training process in Algorithm~\ref{algo:ecg_byte} and pseudocode of the \texttt{merge} and \texttt{get\_stats} functions in Appendix~\ref{apd:pseudo}.

\begin{algorithm}[t]
\caption{Training Process for \textbf{ECG-Byte}}
{\bfseries Input:} Input $X_{\text{concat}}$, number of merges \texttt{num\_merges}. \par
{\bfseries Output:} Tuple containing final encoded IDs, vocabulary map, and merge history. \par
\begin{algorithmic}[1]
    \STATE \texttt{ids} $\gets$ Convert $X_{\text{concat}}$ to a vector of $u32$
    \STATE \texttt{vocab} $\gets$ Mappings from IDs $0$ to $255$ as string\\
    \STATE \texttt{vocab\_tokens} $\gets$ Mappings from IDs $0$ to $255$ to singleton lists
    \STATE \texttt{merges} $\gets$ Empty list
    \FOR{$i \gets 0$ \textbf{to} \texttt{num\_merges} - 1}
        \STATE \texttt{pairs} $\gets$ \textit{get\_stats}(\texttt{ids})
        \IF{\texttt{pairs} is empty}
            \STATE \textbf{break}
        \ENDIF
        \STATE \texttt{best\_pair} $\gets$ pair in \texttt{pairs} with highest frequency
        \IF{\texttt{best\_pair} is not found}
            \STATE \textbf{break}
        \ENDIF
        \STATE \texttt{new\_id} $\gets 256 + i$
        \STATE \texttt{ids} $\gets$ \textit{merge}(\texttt{ids}, \texttt{best\_pair}, \texttt{new\_id})
        \STATE \texttt{vocab}[\texttt{new\_id}] $\gets$ \\\textit{concat}(\texttt{vocab}[\texttt{best\_pair.0}], \texttt{vocab}[\texttt{best\_pair.1}])
        \STATE \texttt{new\_token} $\gets$ \\\textit{concat}(\texttt{vocab\_tokens}[\texttt{best\_pair.0}], \texttt{vocab\_tokens}[\texttt{best\_pair.1}])
        \STATE \texttt{vocab\_tokens}[\texttt{new\_id}] $\gets$ \texttt{new\_token}
        \STATE Append (\texttt{new\_token}, \texttt{new\_id}) to \texttt{merges}
    \ENDFOR
    \STATE \textbf{return} (\texttt{ids}, \texttt{vocab}, \texttt{merges})
\end{algorithmic}
\label{algo:ecg_byte}
\end{algorithm}

\paragraph{ECG-Byte Encoding Process} After training \textbf{ECG-Byte}, we encode any quantized ECG signal $X_{\text{symb}}$ by first converting each byte to a 32-bit unsigned integer and building a trie structure, where each node represents a byte or a merged token sequence from prior encoding steps.
The trie is initialized with single-byte tokens (0-255) and is extended with custom token sequences from the learned merge history. 
For each byte sequence in the input, the encoding function traverses the trie to find the longest match, replacing matched sequences with their assigned token IDs. 
If no match is found, the byte is added to the output as-is.
The final encoded sequence is returned as \texttt{output\_ids}, where we will denote as $X_{\text{ID}}$.
We provide the detailed pseudocode of the encoding process in Algorithm~\ref{algo:ecg_byte_encode} in the Appendix due to page limitations.

\subsection{Large Language Model}
In this study, we utilize the Llama-3.2-1B \citep{grattafiori2024llama3herdmodels} checkpoint through the HuggingFace API \citep{wolf2020huggingfaces} unless specified otherwise.
We also provide an ablation study in subsection~\ref{ablstudy} where we utilize other popular LLMs, such as GPT2 XL 1.5B \citep{radford_2019_language}, Gemma 2B \citep{gemmateam2024gemma2improvingopen}, and OPT 1.3B \citep{zhang2022optopenpretrainedtransformer}. 
With the exception of modifying the token embedding size (since we add new ECG tokens), we utilize all default hyperparameters out-of-the-box unless specified otherwise.

\subsection{Learning Objective}
The learning objective for training the LLM considers a sequence composed of three parts, \( \{ X_{\text{ID}}, Q, \mathcal{S} \} \), where \( X_{\text{ID}} \in \mathcal{V}^{M} \) represents the encoded ECG sequence of length \( l_{X_{\text{ID}}} = |X_{\text{ID}}| \), with each token drawn from the extended vocabulary \( \mathcal{V} \) of size \( M \), \( Q \) represents the tokenized question, and \( \mathcal{S} \) denotes the tokenized answer sequence. 
The input sequence includes special tokens: \( \texttt{[BOS]} \) as the beginning-of-sequence token, \( \texttt{[SIG\_START]} \) and \( \texttt{[SIG\_END]} \) to indicate the start and end of the encoded ECG sequence, and \( \texttt{[EOS]} \) as the end-of-sequence token for the generated answer.
The motivation for adding \( \texttt{[SIG\_START]} \) and \( \texttt{[SIG\_END]} \) special tokens is inspired by \citet{liu2023visualinstructiontuning}, where they utilize special tokens indicating the start and end of the image.
Thus, the full input sequence is structured as:
$\texttt{[BOS]} \| \texttt{[SIG\_START]} \| X_{\text{ID}} \| \texttt{[SIG\_END]} \| Q \| \mathcal{S} \| \texttt{[EOS]},$
where \( \| \) denotes concatenation.
Let \( l_Q = |Q| \), \( l_{\mathcal{S}} = |\mathcal{S}| \), and \( L \) be the total sequence length, given by $L = 4 + l_{X_{\text{ID}}} + l_Q + l_{\mathcal{S}},$ where 4 is accounting for the \( \texttt{[BOS]}, \texttt{[SIG\_START]}, \texttt{[SIG\_END]}, \) and \( \texttt{[EOS]} \) tokens. 
The autoregressive objective maximizes the likelihood of each token in \( \mathcal{S} \| \texttt{[EOS]} \) conditioned on the preceding context \( \texttt{Context} = \{ \texttt{[BOS]}, \texttt{[SIG\_START]}, X_{\text{ID}}, \texttt{[SIG\_END]}, Q \} \) and the previous tokens in \( \mathcal{S} \). 
The objective is formulated as follows:
\begin{equation}
    \mathcal{L}_{\text{NLL}} = - \sum_{l^` = l_{X_{\text{ID}}} + l_Q + 4}^{L} \log P(s_{l^`} \mid \texttt{Context}, s_{<{l^`}}; \theta),
\end{equation}
where \( s_{l^`} = \mathcal{S}_{{l^`} - (l_{X_{\text{ID}}} + l_Q + 4)} \) is the \(({l^\prime} - (l_{X_{\text{ID}}} + l_Q + 4))\)-th token in \( \mathcal{S} \| \texttt{[EOS]} \), and \( s_{<{l^`}} = \{s_1, s_2, \dots, s_{{l^`} - (l_{X_{\text{ID}}} + l_Q + 4) - 1}\} \) denotes all tokens in \( \mathcal{S} \| \texttt{[EOS]} \) preceding \( s_{l^`} \).

\section{Experiments}
\subsection{Experimental Settings}
We finetuned the LLM using the Adam optimizer \citep{kingma2017adam} with a learning rate of $1e-4$, weight decay of $1e-2$ \citep{NIPS1991_8eefcfdf}, and a custom learning rate scheduler. 
We provide details about the learning rate scheduler in Appendix~\ref{lr_sched} due to page limitations.
We train the LLMs for 1 epoch and with a batch size of 2.
We set the exponential decay rates to be $\beta_1 = 0.9$ and $\beta_2 = 0.99$.
We set $\epsilon = 1e-8$ as a constant for numerical stability.
Due to computational constraints, we limit each dataset to a randomly sampled training subset of 400,000 ECG instances and an independent inference subset of 25,000 instances, unless specified otherwise.
We also utilize LoRA \citep{hu2021loralowrankadaptationlarge} to finetune the LLM with rank = 16, $\alpha_{LoRA} = 32$, and dropout = 0.05.
We conduct our experiments on 4 NVIDIA RTX A6000 48 GB GPUs.

During inference, we evaluate our model with number of merges \texttt{num\_merges} = 3500, sequence length $L = 1024$, and ECG length $T=500$ unless specified otherwise. We use popular metrics for NLG namely the BLEU-4 \citep{Papineni2002BleuAM}, Rouge-L \citep{Lin2004ROUGEAP}, Meteor \citep{Banerjee2005METEORAA}, and BertScore F1 \citep{Zhang2020BERTScoreET} metrics.
All reported NLG results are presented as means with standard deviations computed over 5 random seeds.

\begin{table*}[hbtp]
  \caption{NLG mean results with standard deviations over 5 random seeds comparing against different baselines.}
  {\resizebox{\textwidth}{!}{%
  \small \begin{tabular}{lllcccc}
  \toprule
  \bfseries Method & \bfseries Trained Dataset & \bfseries Inferenced Dataset & \bfseries BLEU-4 & \bfseries Rouge-L & \bfseries Meteor & \bfseries BertScore F1 \\
  \midrule
  ECG-Chat \citep{zhao2024ecgchatlargeecglanguagemodel} & \multirow{6}{*}{MIMIC-IV ECG Pretrain} &\multirow{6}{*}{PTB-XL} & \textbf{11.19} & 29.93 & \textbf{35.10 }& - \\
  $L_{CL}$ &  & & 8.10 ± 0.25 & 31.36 ± 0.31 & 27.55 ± 0.36 & 89.35 ± 0.04 \\
  $L_{MIM}$ &  & & 6.21 ± 0.22 & 30.63 ± 0.13 & 24.91 ± 0.14 & \textbf{90.44} ± 0.04 \\
  $L_{MERL}$ \citep{liu2024zeroshotecgclassificationmultimodal} &  & & 10.22 ± 0.25 & 32.95 ± 0.12 & 25.60 ± 0.17 & 89.94 ± 0.01 \\
  $L_{Dual}$ & & & 9.33 ± 0.22  & 30.45 ± 0.21 & 24.37 ± 0.36 & 90.29 ± 0.02 \\
  \textbf{ECG-Byte} & & & 11.00 ± 0.19 & \textbf{33.41} ± 0.05& 24.95 ± 0.09 & 90.02 ± 0.01 \\
  \midrule
  $L_{CL}$ & \multirow{5}{*}{ECG-QA MIMIC-IV} & \multirow{5}{*}{ECG-QA MIMIC-IV} & 10.22 ± 0.06 & 38.41 ± 0.48 & 24.66 ± 0.23& 90.42 ± 0.09 \\
  $L_{MIM}$ &  & & 7.90 ± 0.23 & 29.28 ± 0.38 & 19.03 ± 0.11 & 67.91 ± 0.17 \\
  $L_{MERL}$ \citep{liu2024zeroshotecgclassificationmultimodal}&  && 10.95 ± 0.24& 38.18 ± 0.58 & 26.24 ± 0.36 & 90.80 ± 0.06 \\
  $L_{Dual}$ &  && 8.57 ± 0.14 & 34.00 ± 0.25 & 25.22 ± 0.30 & 87.72 ± 0.04 \\
  \textbf{ECG-Byte} & & & \textbf{11.23} ± 0.12 & \textbf{42.49} ± 0.53 & \textbf{27.08} ± 0.15 & \textbf{91.30} ± 0.04 \\
  \midrule
  $L_{CL}$& \multirow{5}{*}{ECG-QA PTB-XL} & \multirow{5}{*}{ECG-QA PTB-XL} & 8.89 ± 0.25 & 28.63 ± 0.47 & 18.45 ± 0.31 & 72.63 ± 0.40 \\
  $L_{MIM}$ & & & \textbf{15.14} ± 0.28 & 46.71 ± 0.41 & \textbf{29.64} ± 0.30 & 92.12 ± 0.10 \\
  $L_{MERL}$ \citep{liu2024zeroshotecgclassificationmultimodal}& & & 13.84 ± 0.19& 40.14 ± 0.39 & 26.24 ± 0.35 & 91.88 ± 0.09 \\
  $L_{Dual}$ & & & 14.72 ± 0.27& 42.88 ± 0.13 & 28.25 ± 0.27 & 89.40 ± 0.01 \\
  \textbf{ECG-Byte} & & & 13.93 ± 0.21 & \textbf{47.08} ± 0.56 & 29.17 ± 0.31 & \textbf{92.53} ± 0.07 \\
  \bottomrule
  \end{tabular}}}
  \label{tab:main_results}
\end{table*}

\section{Results}
\label{results}
\subsection{Natural Language Generation}
We present our main results in Table~\ref{tab:main_results}, comparing \textbf{ECG-Byte} with prior works and self-implemented two-stage pretraining methods. Notably, \citet{zhao2024ecgchatlargeecglanguagemodel} is not directly comparable due to differing data splits and pretraining datasets. 
\citet{zhao2024ecgchatlargeecglanguagemodel} train on the \textit{full} MIMIC-IV ECG Pretrain dataset, finetune on an instruction-tuning dataset for ECG-related conversations, and evaluate on PTB-XL \citep{wagner_ptb-xl_2020} using a unified question: ``Could you please help me explain my ECG?"
We also note that ECG-Chat \citep{zhao2024ecgchatlargeecglanguagemodel} employs numerous generation enhancements such as Retrieval Augmented Generation (RAG), DSPy for automatic prompt tuning, and an adapted Diagnosis-Driven Prompt (DDP) inspired by \citet{jin2024promptmrgdiagnosisdrivenpromptsmedical}. 
In our work, we only observe performance of NLG given a text prompt and ECG signal. 
Therefore, to establish comparable baselines, we implement generic two-stage pretraining methods utilizing strong encoders: $L_{CL}$, $L_{MIM}$, and $L_{Dual} = L_{CL} + L_{MIM}$. Here, $L_{CL}$ employs contrastive learning \citep{liu2024zeroshotecgclassificationmultimodal, gopal20213kgcontrastivelearning12lead, pham2024cmeltcontrastiveenhancedmasked, kiyasseh2021clocscontrastivelearningcardiac}, $L_{MIM}$ uses Masked Image Modeling (MIM) \citep{choi2023ecgbert, na2024guidingmaskedrepresentationlearning, 9926900}, and $L_{Dual}$ combines both \citep{oh2022leadagnosticselfsupervisedlearninglocal, mckeen2024ecgfmopenelectrocardiogramfoundation}.
All three implementations utilize pretrained CLIP \citep{Radford2021LearningTV} and ViT \citep{dosovitskiy2021image} models, where ECG signals are transformed into three-channel images for finetuning.
Additionally, we adapt \citet{liu2024zeroshotecgclassificationmultimodal}'s state-of-the-art contrastive method ($L_{MERL}$) for fair comparison. 
Their most effective model uses a 1D ResNet backbone \citep{he2015deepresiduallearningimage}, therefore we utilize the ResNet101 variant. 
For $L_{CL}$, $L_{MIM}$, $L_{Dual}$, and $L_{MERL}$, training is conducted on the \textit{full}, preprocessed MIMIC-IV ECG dataset with a batch size of 64 during the first stage to simulate prior work settings. 
Implementation details for both training stages are in Appendix~\ref{apd:second}.
Table~\ref{tab:main_results} demonstrates \textbf{ECG-Byte}'s effectiveness, showing competitive or superior performances across all metrics and datasets compared to other methods. We also highlight that \textbf{ECG-Byte} was initially trained on a separate, unsegmented subset of 200,000 ECGs from the MIMIC-IV ECG dataset. With this fact, \textbf{ECG-Byte} displays strong performances when training and inferencing the LLM on ECGs of different lengths (i.e., segmented to 2 seconds) and from another dataset (i.e., PTB-XL). Qualitative examples of generated answers are provided in Appendix~\ref{apd:nlg}.

\subsection{Efficiency of ECG-Byte}
We compare the efficiency of our end-to-end (end-to-end A/B) approach (\textbf{ECG-Byte} A/B) with two-stage pretraining (1st Stage / 2nd Stage) in Table~\ref{tab:efficiency}. 
The top and middle sections report data and time requirements for both settings used in our study. 
The two-stage methods first train on the full MIMIC-IV ECG dataset \citep{johnson_2023_mimiciv} using segmented ECGs, which are then used as inputs in the second stage. 
Although \textbf{ECG-Byte} is trained on unsegmented ECGs, we convert these counts to segmented-equivalents and reduce data requirements by sampling only a subset for tokenizer training. 
Training time for two-stage methods is averaged over our self-implemented approaches ($L_{CL}$, $L_{MIM}$, $L_{Dual}$) and the $L_{MERL}$ method \citep{liu2024zeroshotecgclassificationmultimodal}. 
Under these settings, our method achieves competitive results using approximately \textbf{48\% of the data} and is about \textbf{3 times faster} in total training time. 
Even when training \textbf{ECG-Byte} on 500,000 unsegmented ECGs (equivalent to 2,500,000 segmented ECGs) as seen in the bottom section (\textbf{ECG-Byte} B), our method remains \textbf{2.38 times faster} than two-stage pretraining, highlighting its efficiency.

\begin{table}[!htp]
\centering
\caption{Efficiency of our method compared against two-stage pretraining methods.}
{\resizebox{0.8\columnwidth}{!}{%
\begin{tabular}{lrrrrr}\toprule
\textbf{Method} &\textbf{\# of Data} &\textbf{Total \# of Data} &\textbf{Time (min.)} &\textbf{Total Time (min.)} \\\midrule
1st Stage &2,513,435 &\multirow{2}{*}{2,913,435} &$\sim$1258.50 &\multirow{2}{*}{$\sim$1727.75} \\
2nd Stage &400,000 & &$\sim$469.25 & \\
\midrule
\textbf{ECG-Byte} A &1,000,000 &\multirow{2}{*}{1,400,000} & $\sim$152.47&\multirow{2}{*}{$\sim$\textbf{572.79}} \\
end-to-end A &400,000 & &$\sim$420.32 & \\
\midrule
\textbf{ECG-Byte} B &2,500,000 &\multirow{2}{*}{2,900,000} &$\sim$304.27 &\multirow{2}{*}{$\sim$\textbf{724.59}} \\
end-to-end B &400,000 & &$\sim$420.32 & \\
\bottomrule
\end{tabular}}}
\label{tab:efficiency}
\end{table}

\begin{table*}[hbtp]
  \caption{Mean results with standard deviations over 5 random seeds on training two-stage pretraining methods in an end-to-end fashion.}
  {\resizebox{\textwidth}{!}{%
  \small \begin{tabular}{lllcccc}
  \toprule
  \bfseries Method & \bfseries Trained Dataset & \bfseries Inferenced Dataset& \bfseries BLEU-4 & \bfseries Rouge-L & \bfseries Meteor & \bfseries BertScore F1 \\
  \midrule
  $L_{MIM}$ & \multirow{5}{*}{ECG-QA MIMIC-IV} &\multirow{5}{*}{ECG-QA MIMIC-IV} & 7.90 ± 0.23 & 29.28 ± 0.38 & 19.03 ± 0.11 & 67.91 ± 0.17 \\
  End-to-End $L_{MIM}$ &  &  & 7.71 ± 0.36 & 28.19 ± 0.13 & 18.88 ± 0.54 & 67.82 ± 0.04 \\
  $L_{MERL}$ \citep{liu2024zeroshotecgclassificationmultimodal}&  && 10.95 ± 0.24& 38.18 ± 0.58 & 26.24 ± 0.36 & 90.80 ± 0.06 \\
  End-to-End $L_{MERL}$ \citep{liu2024zeroshotecgclassificationmultimodal}& & & 8.26 ± 0.17 & 34.88 ± 0.26 & 22.14 ± 0.33 & 86.64 ± 0.03 \\
  \textbf{ECG-Byte} & & & \textbf{11.23} ± 0.12 & \textbf{42.49} ± 0.53 & \textbf{27.08} ± 0.15 & \textbf{91.30} ± 0.04 \\
  \midrule
  $L_{MIM}$ & \multirow{5}{*}{ECG-QA PTB-XL}& \multirow{5}{*}{ECG-QA PTB-XL}& \textbf{15.14} ± 0.28 & 46.71 ± 0.41 & \textbf{29.64} ± 0.30 & 92.12 ± 0.10 \\
  End-to-End $L_{MIM}$ &  &  & 12.04 ± 0.17 & 40.44 ± 0.25 & 24.84 ± 0.43 & 87.71 ± 0.11 \\
  $L_{MERL}$ \citep{liu2024zeroshotecgclassificationmultimodal}& & & 13.84 ± 0.19& 40.14 ± 0.39 & 26.24 ± 0.35 & 91.88 ± 0.09 \\
  End-to-End $L_{MERL}$ \citep{liu2024zeroshotecgclassificationmultimodal}&  &  &11.27 ± 0.34 & 38.83 ± 0.17 & 24.03 ± 0.21 & 85.64 ± 0.32  \\
  \textbf{ECG-Byte} & & & 13.93 ± 0.21 & \textbf{47.08} ± 0.56 & 29.17 ± 0.31 & \textbf{92.53} ± 0.07 \\
  \bottomrule
  \end{tabular}}}
  \label{tab:both}
\end{table*}

\subsection{Training Two-Stage Pretraining Methods in an End-to-End Fashion}
Although training methods are conventionally categorized as either two-stage or end-to-end, some works adopt an end-to-end strategy using an ECG-specific encoder architecture typically seen in two-stage methods \citep{wan2024meitmultimodalelectrocardiograminstruction}. In Table~\ref{tab:both}, we simulate the approach of \citet{wan2024meitmultimodalelectrocardiograminstruction} by jointly training the ECG encoder, learnable projection matrix, and LLM end-to-end using only an autoregressive objective. This contrasts with the two-stage method, where the ECG encoder is first trained with a self-supervised learning (SSL) objective and then frozen. We select strong ECG encoders from Table~\ref{tab:main_results}—namely, $L_{MIM}$ and $L_{MERL}$—and report performance on the ECG-QA MIMIC-IV and PTB-XL datasets \citep{oh2023ecgqacomprehensivequestionanswering}. We denote the traditional 2-stage training approach for both ECG encoders as $L_{MIM}$ and $L_{MERL}$ and the end-to-end adaptation of 2-stage training approaches as End-to-End $L_{MIM}$ and End-to-End $L_{MERL}$ respectively in Table~\ref{tab:both}. The results clearly show a performance drop when the ECG encoder is not pretrained on a large dataset. This finding aligns with the Vision-Language Model (VLM) domain, where vision encoders are initially trained on internet-scale image data before integration with an LLM for visual understanding \citep{liu2023visualinstructiontuning}. Similarly, these findings indicate that ECG signals require extensive first-stage training with a dedicated self-supervised learning objective to learn robust representations before being applied to NLG. Even powerful, general-purpose pretrained encoders like ViT \citep{dosovitskiy2021image} necessitate an initial training phase to effectively capture the unique characteristics of ECG data. In contrast, our method, \textbf{ECG-Byte}, compresses ECG signals using a rule-based algorithm (e.g., BPE) to represent them as tokens, enabling direct LLM training. Although these tokens belong to a different modality, we believe that the similarity in input representation allows the LLM to adapt more effectively without extensive first-stage training.

\subsection{Ablation Study}
\label{ablstudy}
We conduct several ablation studies to show the variability of performance with \textbf{ECG-Byte} when we use varying LLMs during finetuning, sequence lengths $L$, and ECG lengths $T$.
With the exception of the ablating parameter, we fix all other parameters to Llama 3.2 1B, \texttt{num\_merges} = 3500, $L = 1024$, and $T=500$.
We report results when training and testing on the PTB-XL variant of ECG-QA \citep{oh2023ecgqacomprehensivequestionanswering} unless specified otherwise.

\begin{table}[!hbtp]
  \caption{Ablation study on using different LLMs.}
  {\resizebox{\columnwidth}{!}{%
  \small \begin{tabular}{lcccc}
  \toprule
  \bfseries LLM & \bfseries BLEU-4 & \bfseries Rouge-L & \bfseries Meteor & \bfseries BertScore F1 \\
  \midrule
  GPT2 XL 1.5B \citep{radford_2019_language}& 12.30 ± 0.19 & 41.33 ± 0.57 & 26.48 ± 0.33 & 92.00 ± 0.06 \\
  Gemma 2B \citep{gemmateam2024gemma2improvingopen}& 13.78 ± 0.18 & 45.48 ± 0.55 & 28.32 ± 0.23 & 92.01 ± 0.02 \\
  OPT 1.3B \citep{zhang2022optopenpretrainedtransformer}& 12.26 ± 0.20 & 41.84 ± 0.52 & 26.21 ± 0.29 & 91.78 ± 0.04 \\
  Llama 3.2 1B \citep{grattafiori2024llama3herdmodels}& \textbf{13.93} ± 0.21 & \textbf{47.08} ± 0.56 & \textbf{29.17} ± 0.31 & \textbf{92.53} ± 0.07 \\
  \bottomrule
  \end{tabular}}}
  \label{tab:diff_llm}
\end{table}

\paragraph{Different LLMs}
We show the variability in performance of \textbf{ECG-Byte} when using different LLMs with similar numbers of parameters in Table~\ref{tab:diff_llm}. 
While Llama 3.2 1B \citep{grattafiori2024llama3herdmodels} achieves the best results, GPT2 XL 1.5B \citep{radford_2019_language}, Gemma 2B \citep{gemmateam2024gemma2improvingopen}, and OPT 1.3B \citep{zhang2022optopenpretrainedtransformer} also deliver comparable performances. 
These findings demonstrate that our method is not limited to Llama 3.2 1B but can achieve similar results across a variety of LLMs. 

\begin{table}[!hbtp]
  \centering
  \begin{minipage}{0.47\textwidth}
    \centering
    \caption{Ablation study on varying sequence lengths \( L \).}
    {\resizebox{\columnwidth}{!}{%
    \small
    \begin{tabular}{lcccc}
      \toprule
      \( L \) & BLEU-4 & Rouge-L & Meteor & BertScore F1 \\
      \midrule
      512  & 13.61 ± 0.15  & \textbf{48.15} ± 0.57 & 29.10 ± 0.28 & 92.41 ± 0.05 \\
      1024 & \textbf{13.93} ± 0.21 & 47.08 ± 0.56     & \textbf{29.17} ± 0.31 & \textbf{92.53} ± 0.07 \\
      2048 & 13.88 ± 0.22  & 45.21 ± 0.48     & 28.31 ± 0.27 & 90.88 ± 0.02 \\
      \bottomrule
    \end{tabular}}}
    \label{tab:seq_len}
  \end{minipage}%
  \hfill
  \begin{minipage}{0.47\textwidth}
    \centering
    \caption{Ablation study on varying lengths \( T \).}
    {\resizebox{\columnwidth}{!}{%
    \small
    \begin{tabular}{lcccc}
      \toprule
      \( T \) & BLEU-4 & Rouge-L & Meteor & BertScore F1 \\
      \midrule
      250  & 12.64 ± 0.20  & 47.31 ± 0.26 & 27.97 ± 0.21 & 92.32 ± 0.06 \\
      500  & 13.93 ± 0.21  & 47.08 ± 0.56 & 29.17 ± 0.31 & 92.53 ± 0.07 \\
      1250 & 11.01 ± 0.19  & 43.84 ± 0.28 & 25.49 ± 0.20 & \textbf{93.07} ± 0.03 \\
      2500 & \textbf{14.54} ± 0.17  & \textbf{48.03} ± 0.27 & \textbf{32.11} ± 0.22 & 92.91 ± 0.04 \\
      \bottomrule
    \end{tabular}}}
    \label{tab:len_ecg}
  \end{minipage}
\end{table}

\begin{figure}[htp]
\centering
\includegraphics[width=0.8\linewidth]{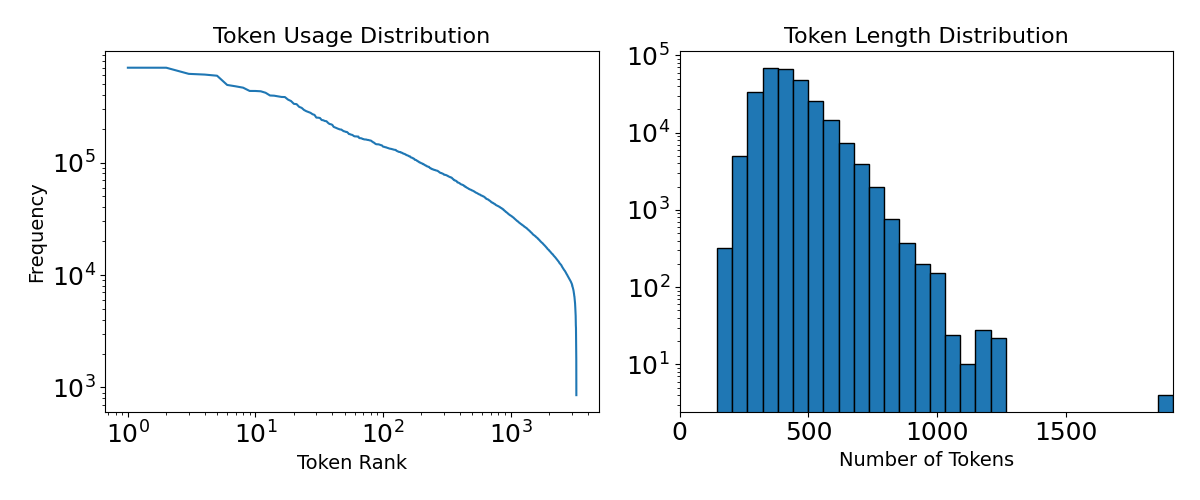}
\caption{Plots of the token usage and length distributions for \textbf{ECG-Byte} where $\texttt{num\_merges} = 3500$. More examples with varying $\texttt{num\_merges}$ are provided in Appendix~\ref{apd:token-dist}.}
\label{Fig:token-dist}
\end{figure}

\paragraph{Sequence Length}
Input lengths for LLMs are critical for efficient training since the computation of attention scales quadratically with sequence length \citep{vaswani2023attention}. Table~\ref{tab:seq_len} presents results for various sequence lengths \( L \), which reveal minimal performance differences. As Figure~\ref{Fig:token-dist} shows, most ECGs are encoded with token lengths around 500. Thus, we hypothesize that even small \( L \) values (e.g., \( L=512 \)) preserve the complete ECG information, resulting in only slight performance variations. We note that this can change depending on the initial ECG length $T$, in which we ablate in the next paragraph. We leave it up to future works for finding the sweet spot for the sequence length $L$ and its interaction with the initial ECG length $T$.

\paragraph{ECG length}
Lastly, we show the effect of the length $T$ being considered when encoding the ECG with \textbf{ECG-Byte} in Table~\ref{tab:len_ecg}.
We want to note that for the results of $T = 2500$, the full unsegmented ECG is utilized. 
Consequently, the number of instances available is less than the targeted dataset size of 400,000 (i.e., 97,244).
Thus, when $T=2500$, we use the full dataset to train the model. 
For shorter segment lengths, such as $T = 250$ and $T = 500$, the model demonstrates strong performances indicating that shorter segments can effectively preserve relevant information for NLG. 
Interestingly, for $T = 2500$, the model achieves the highest performance across all metrics. 
This suggests that when the model is trained with the full 10 second encoded ECG, it benefits from richer contextual information present in the complete ECG waveform.

\subsection{ECG-Byte Analysis}
\label{ecg:an}
We analyze \textbf{ECG-Byte} by visualizing the usage of merged tokens, length of the encoded ECG, and mapping between the encoded tokens and original ECG.
Unless specified otherwise, we analyze \textbf{ECG-Byte} when $\texttt{num\_merges} = 3500$, $L = 1024$, and $T=500$.

\paragraph{Token Usage and Length Distribution}
We examine the token usage and length distributions for \textbf{ECG-Byte} with \texttt{num\_merges} = 3500 on a subsample of 277,840 ECGs from the PTB-XL dataset. 
For a 2-second ECG ($T=500$), the raw signal comprises $12\times500=6000$ symbols. \textbf{ECG-Byte} then compresses these 6000 symbols at an average ratio of 12.66x, yielding approximately 500 tokens per ECG on average, as illustrated in Figure~\ref{Fig:token-dist}. 
The left panel of Figure~\ref{Fig:token-dist} displays the token usage distribution, showing token frequency (y-axis) ranked in descending order (x-axis). 
A small subset of tokens dominates the occurrences, while the rest are infrequently used—a typical characteristic of BPE-based tokenization, where common patterns are compressed into frequent tokens and rare patterns into infrequent ones. 
The right panel of Figure~\ref{Fig:token-dist} illustrates the token length distribution of the encoded ECGs, with most falling between 500 and 1000 tokens, demonstrating \textbf{ECG-Byte}'s effective compression of the original signal. Additional examples of these distributions are provided in Appendix~\ref{apd:token-dist}.

\begin{figure}[htp]
\centering
\includegraphics[width=0.8\linewidth]{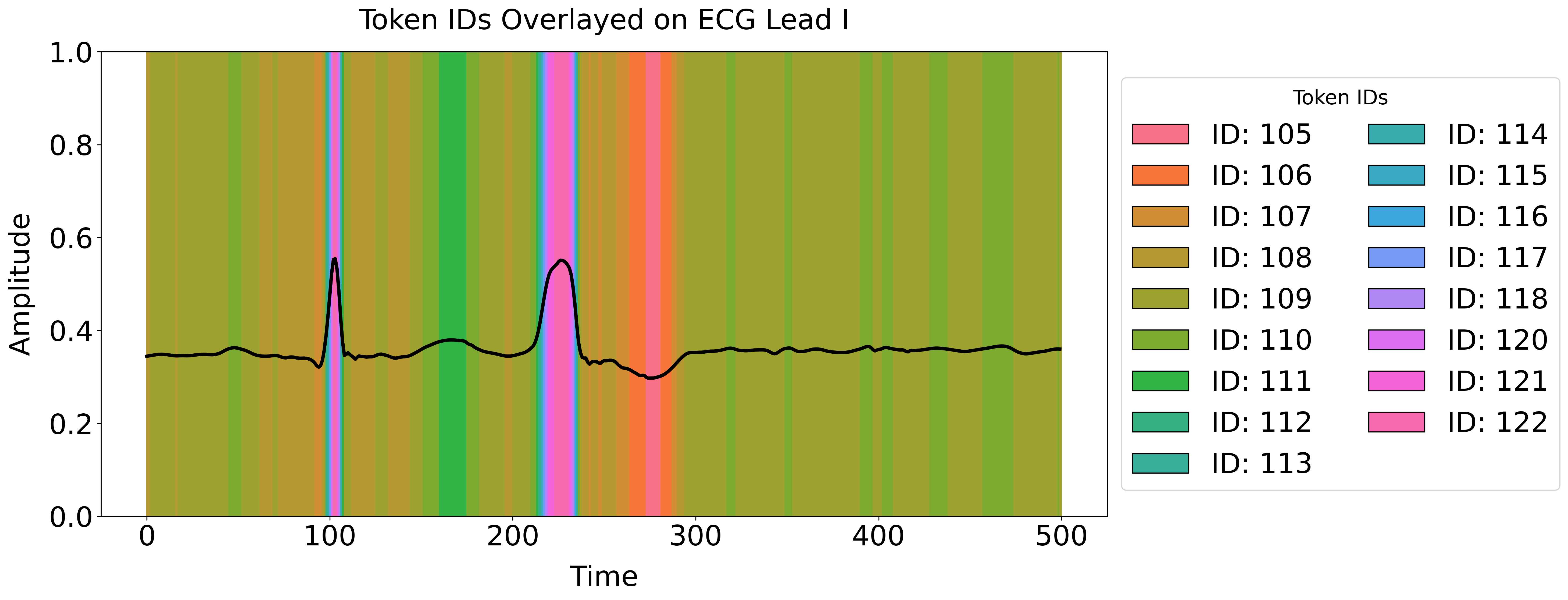}
\caption{A mapping between tokens used for a given ECG Lead I. More examples are provided in Appendix~\ref{apd:map}.}
\label{Fig:ecg-mapping}
\end{figure}

\paragraph{Token to ECG Mapping}
To illustrate how \textbf{ECG-Byte} encodes ECG signals, we analyze the mapping between tokens and signal features. 
Figure~\ref{Fig:ecg-mapping} shows an example Lead I ECG signal with unique token IDs (represented by different colors) overlaid. 
The P wave, QRS complex, and T wave are distinctly captured by different tokens, though this precision varies across instances. 
As demonstrated, \textbf{ECG-Byte} effectively merges key regions of the signal. 
Additional examples are provided in Appendix~\ref{apd:map} due to page limitations.

\begin{figure}[htp]
\centering
\includegraphics[width=0.7\linewidth]{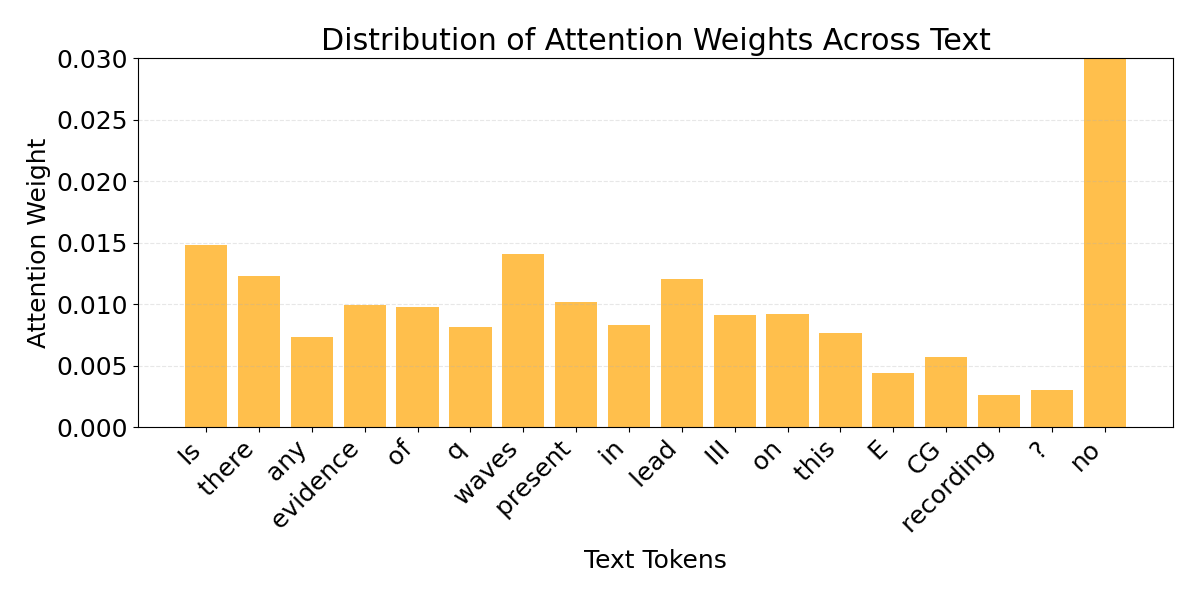}
\includegraphics[width=0.7\linewidth]{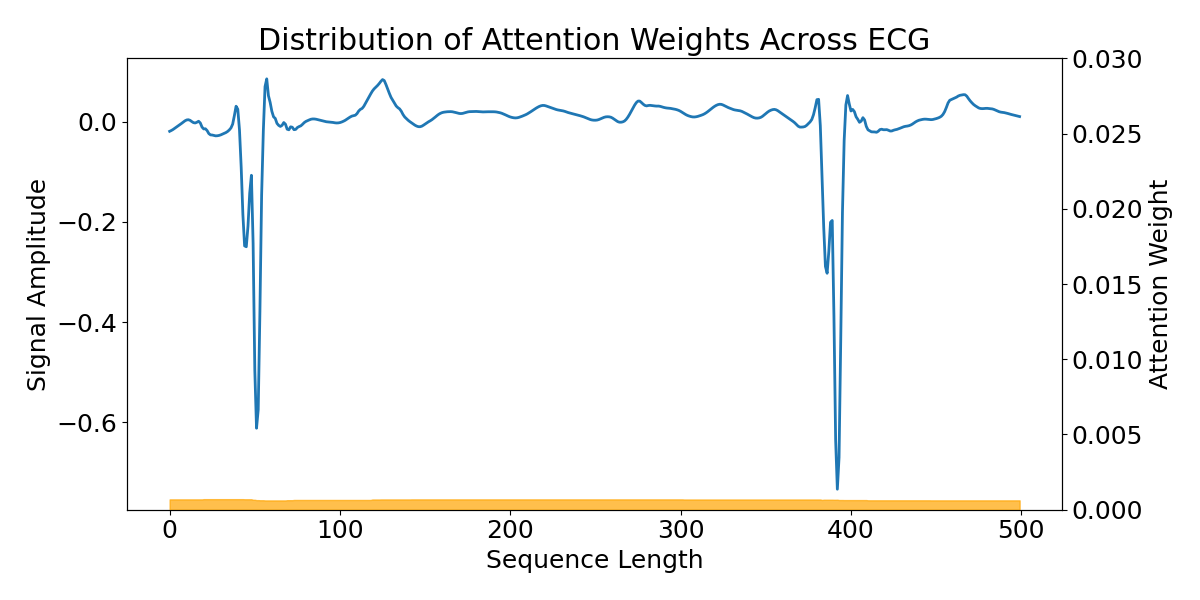}
\caption{The attention weight overlaid on top of both the text (top) and ECG (bottom). More examples are provided in Appendix~\ref{apd:att}.}
\label{Fig:att-mapping2}
\end{figure}

\paragraph{Attention Visualizations}
Figure~\ref{Fig:att-mapping2} visualizes attention weights across a selected ECG lead and text portions of the input after training. 
We focus on one lead due to the uniformity of attention across encoded signal tokens. 
For interpretability, the reversed ECG signal is overlaid on the encoded ECG. 
The model primarily attends to the textual portion of the input sequence, as shown in Figure~\ref{Fig:att-mapping2}.
Previous studies have debated whether attention visualizations are inherently explainable \citep{jain2019attention, wiegreffe2019attention} and explored their role in vision-language models \citep{aflalo2022vlinterpretinteractivevisualizationtool, woo2024dontmissforesttrees, arif2024hiredattentionguidedtokendropping, cui_2024_fading}. 
These works often observe minimal attention to visual input, with models relying primarily on text. 
We hypothesize that a similar phenomenon occurs in Figure~\ref{Fig:att-mapping2}, as the ECG tokens, though represented like text, are 1) newly introduced and 2) perceived as a different modality (e.g., vision). 
We note that attention visualizations may not inherently indicate which parts of the input sequence contribute to the final generated output. Although methods such as LIME \citep{ribeiro2016whyitrustyou} or Integrated Gradients \citep{sundararajan2017axiomaticattributiondeepnetworks} may provide better attributions, we leave this investigation for future work.
Additional examples are provided in Appendix~\ref{apd:att}.

\section{Discussion and Conclusion}
In this study, we introduce \textbf{ECG-Byte}, a custom BPE algorithm to encode ECGs into a discrete sequence of tokens for conditional autoregressive NLG.
\textbf{ECG-Byte} introduces a paradigm shift in generative ECG language modeling by enabling efficient end-to-end training, compared to traditional two-stage pretraining approaches.
Our pipeline demonstrates strong performance, achieving results comparable to two-stage methods while being about \textbf{3 times faster} in total training time and requiring approximately \textbf{48\% of the data}. 
Drawing inspiration from MEIT \citep{wan2024meitmultimodalelectrocardiograminstruction}, we simulate two-stage pretraining for NLG in an end-to-end manner. Our findings indicate that the initial training stage is essential for achieving performance gains. 
In addition to its efficiency, \textbf{ECG-Byte} enhances interpretability.
By analyzing its underlying mechanism, we observe that critical ECG regions, such as the P wave, the QRS complex, and the T wave, are effectively grouped during tokenization, as illustrated in Figure~\ref{Fig:ecg-mapping}.
Furthermore, the reversibility of the compressed token sequence allows us to trace each token back to its original ECG signal segment, providing insight into the specific portions of the signal attended to by the model. 
However, as shown in Figure~\ref{Fig:att-mapping2}, the model’s attention weight distribution resembles that of vision language models, focusing primarily on the textual components of the input sequence during generation.

We emphasize that the goal of \textbf{ECG-Byte} is not to claim it as the best method for representing ECG data or training generative, autoregressive Electrocardiogram-Language Models (ELMs). Instead, \textbf{ECG-Byte} demonstrates the potential of using a \textit{rule-based compressor} rather than a learnable one for ECG data in the context of ELMs. As stated earlier, we find \textbf{ECG-Byte} both highly efficient and intuitive in compressing ECG data, and its tokenization is interpretable because it can be reversed to recover the original signal.
We hope the broader research community will recognize the potential of rule-based compressors for generative tasks with different health signal data beyond ECGs. 
This early-stage work invites the community to contribute to advancing generative ECG language modeling by highlighting some future directions. Future directions include: (1) a comprehensive benchmark with different input representations and training methods for ELMs, (2) refining BPE merging rules to better capture ECG-specific features, (3) adopting more advanced quantization techniques that preserve time-series characteristics \citep{carson2024quantizedsymbolictimeseries, elsworth2020abbaadaptivebrownianbridgebased, carson2024llmabbaunderstandtimeseries}, and (4) introducing stronger modality-specific distinctions, such as embeddings beyond \texttt{[SIG\_START]} and \texttt{[SIG\_END]} \citep{gui2023trainingvisionlanguagetransformerscaptions}.

\paragraph{Limitations}
One key limitation of this work is the scale of both computing resources and data. 
Due to limited computing capacity, we trained on only a data subset using a batch size of 2 for a single epoch, which may have constrained the model's full potential. 
Nevertheless, as prior work indicates that finetuned LLMs can perform satisfactorily with little data \citep{brown2020languagemodelsfewshotlearners}, the promising results of \textbf{ECG-Byte}—in comparison with other two-stage SSL pretraining methods—suggest that this limitation is not a critical bottleneck.

\acks{This work is done in collaboration with the Mario Lemieux Center for Heart Rhythm Care at Allegheny General Hospital. We thank Wenhao Ding, Haohong Lin, Shiqi Liu, Hyoeun Kang, and Tony Chen for the valuable discussions.}

\bibliography{sample}

\newpage
\appendix
\section{Training Details}
\subsection{Learning Rate Scheduler}\label{lr_sched}
We utilize a custom learning rate scheduler in all of our experiments.
This scheduler applies an initial learning rate $\mathrm{init\_lr}$ scaled by the model’s hidden dimension ($\mathrm{d_{model}}^{-0.5}$) and dynamically adjusts it based on training steps, with a warm-up phase of 500 steps.
The learning rate at step $\mathrm{n_{steps}}$ is updated as
$\text{lr} = \mathrm{init\_lr} \times \min \left( \mathrm{n_{steps}}^{-0.5}, \ \mathrm{n_{warmup}}^{-1.5} \times \mathrm{n_{steps}} \right)$.

\section{Additional Details Regarding ECG-Byte}\label{apd:first}

\begin{figure}[htp]
\centering
\includegraphics[width=1\linewidth]{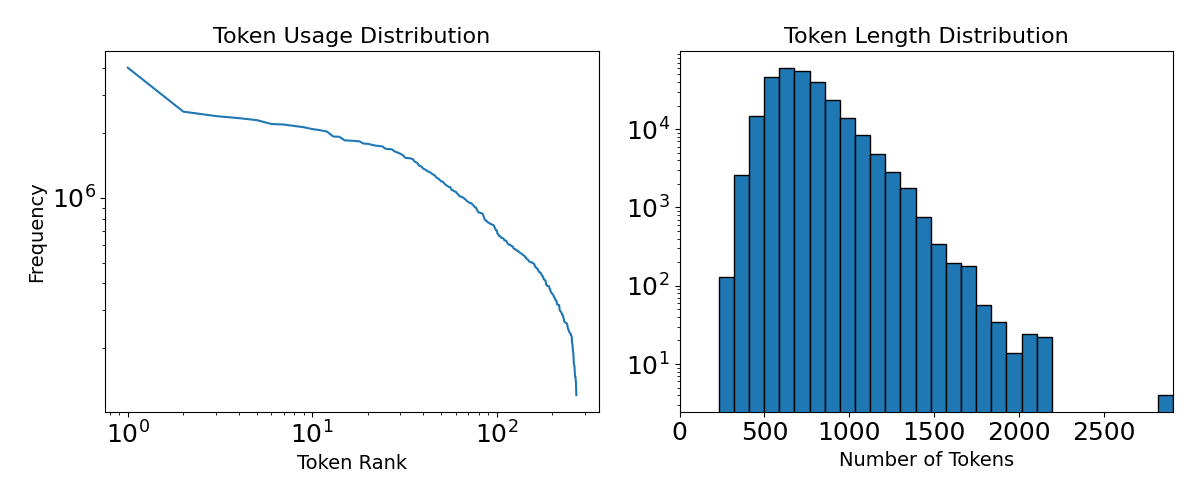}
\includegraphics[width=1\linewidth]{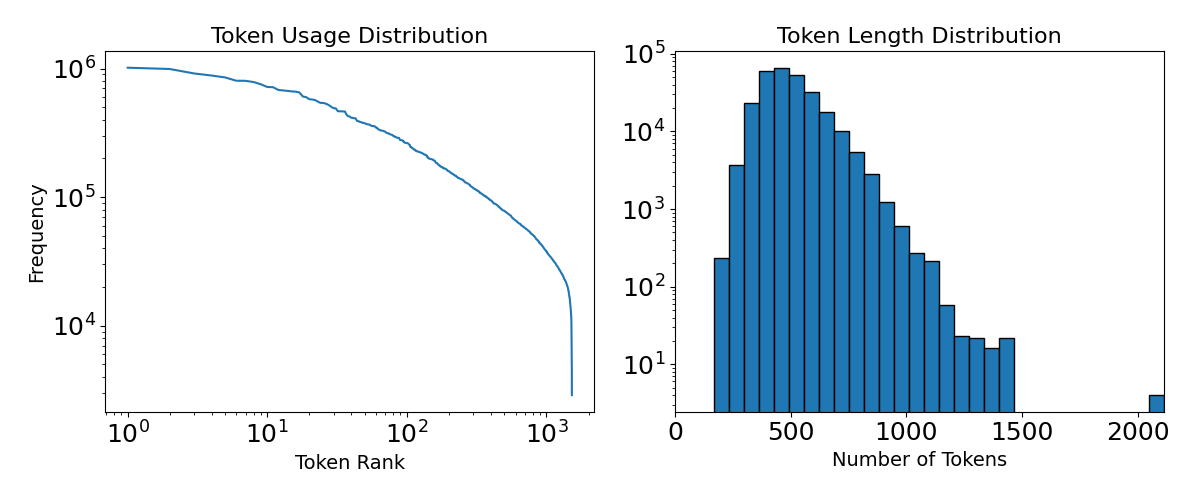}
\includegraphics[width=1\linewidth]{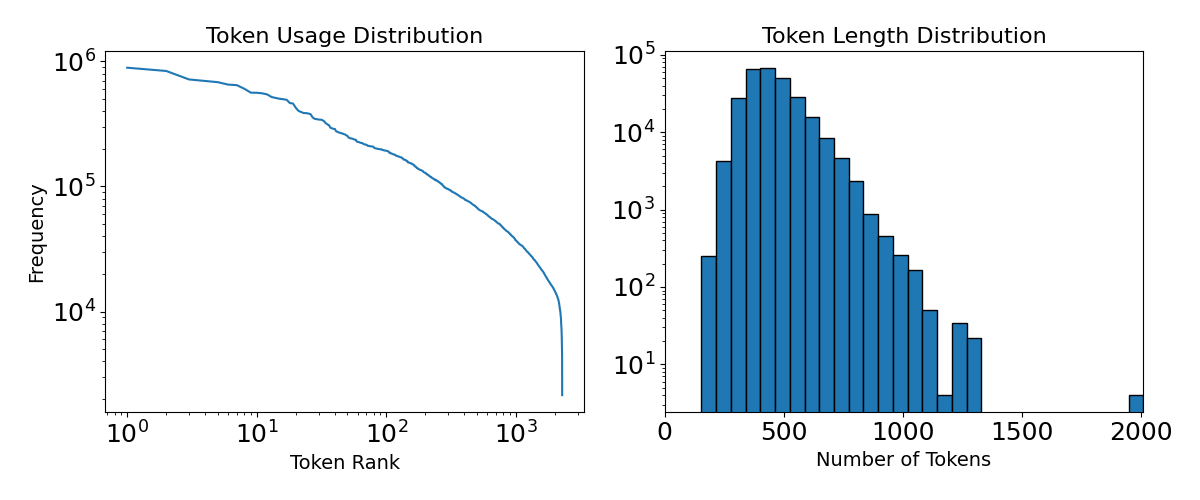}
\caption{Plots of the token usage and length distributions for \textbf{ECG-Byte} where $\texttt{num\_merges}$ is 500, 1750, and 2500 from top to bottom.}
\label{Fig:token-dist-all}
\end{figure}

\subsection{Additional Pseudocode for ECG-Byte}\label{apd:pseudo}
We provide detailed pseudocode for the \textbf{ECG-Byte} encoding process, \texttt{merge} and \texttt{get\_stats} functions in Algorithms~\ref{algo:ecg_byte_encode}, \ref{alg:merge}, and \ref{alg:getstats} respectively.

\subsection{Mapping between Token and ECG}\label{apd:map}
\begin{figure*}[htp]
\centering
\includegraphics[width=1\linewidth]{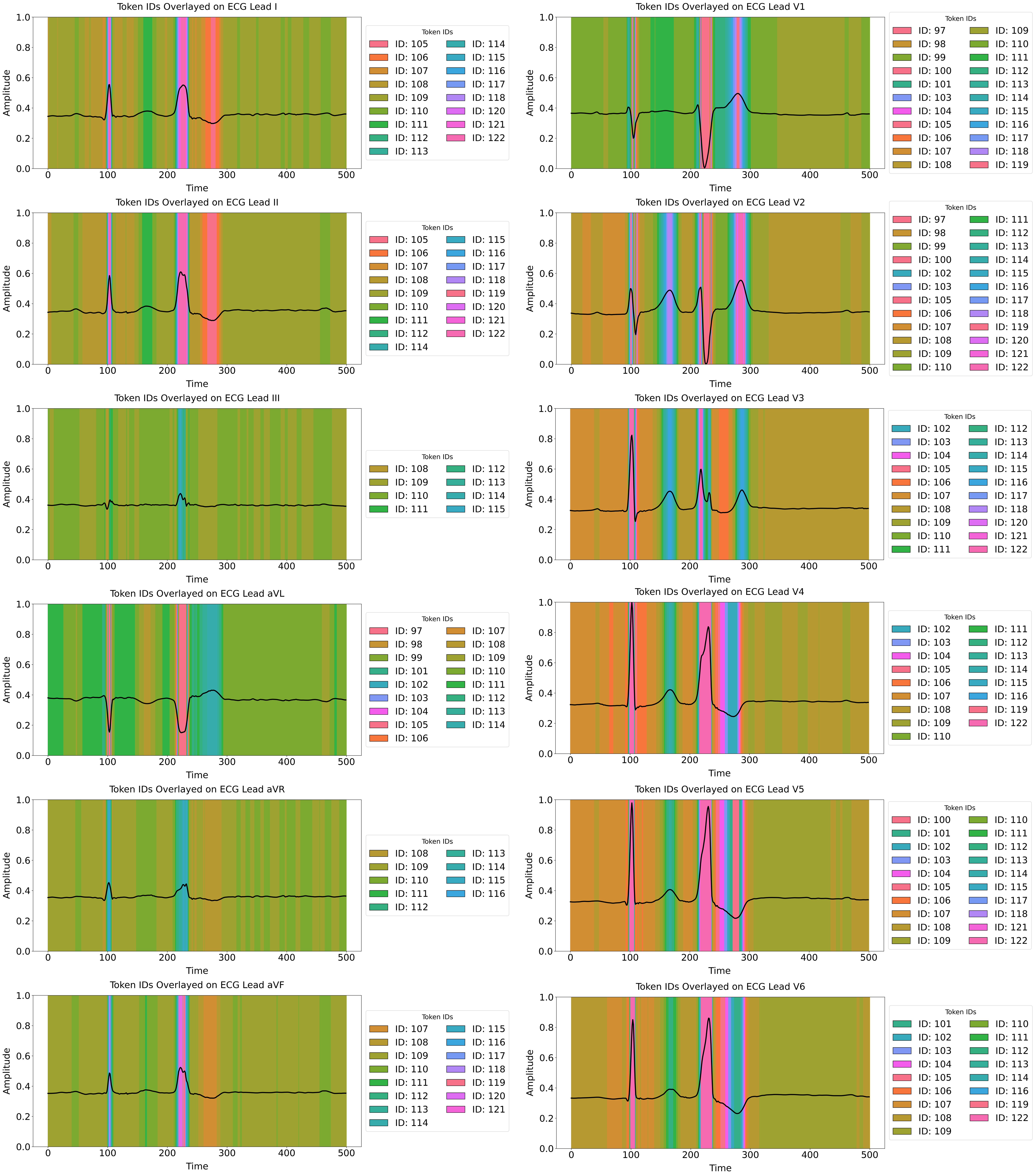}
\caption{A mapping between tokens used for a given ECG Leads I, II, III, aVL, aVR, aVF, V1, V2, V3, V4, V5, V6..}
\label{Fig:ecg-mapping-all}
\end{figure*}

We add more examples of the mapping between the ECG signal and the encoded tokens for \textbf{ECG-Byte} in Figure~\ref{Fig:ecg-mapping-all}.

\subsection{Token usage and length distribution for varying \texttt{num\_merges}}\label{apd:token-dist}
We add more examples of the token usage and length distributions for varying $\texttt{num\_merges}$ in Figure~\ref{Fig:token-dist-all}.

\subsection{Attention Visualizations}\label{apd:att}
\begin{figure*}[htp]
\centering
\includegraphics[width=1\linewidth]{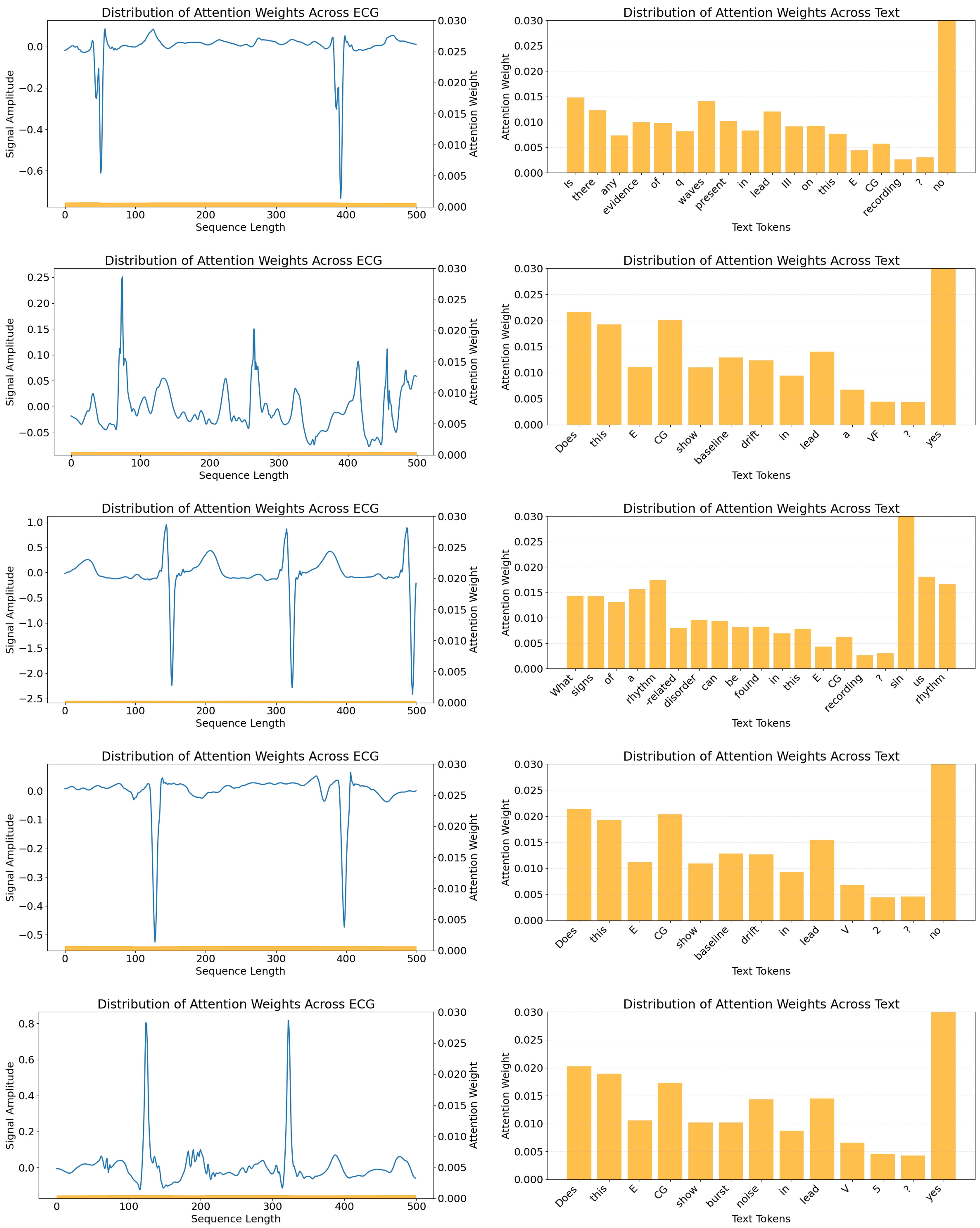}
\caption{The attention weight overlaid on both ECG (left) and text (right).}
\label{Fig:att-all3}
\end{figure*}

We add more visualizations of the attention weights in Figure~\ref{Fig:att-all3}.

\section{Two-stage Pretraining Approaches}\label{apd:second}
To be consistent, we normalize each ECG in the same manner as described in subsection~\ref{ecg-byte-exp}.
Consider a dataset of $N$ ECG-image and clinical note pairs, denoted as $\{(I_i, O_i)\}_{i=1}^N$, where:
$I_i \in \mathbb{R}^{3 \times C \times T}$ is the $i$-th normalized and replicated ECG image, obtained by stacking the clipped ECG signal $X_{\text{clipped}}$ along the channel dimension: $I_i = \text{stack}(X_{\text{clipped}}, X_{\text{clipped}}, X_{\text{clipped}})$.
The reason we do this is because we need to create RGB images to use pretrained image models like ViT \citep{dosovitskiy2021image} and CLIP \citep{Radford2021LearningTV}.

$O_i$ is the corresponding clinical note for the $i$-th ECG, serving as the textual description.
Note that $O_i$ differs from $S$ in the autoregressive setup, where $S$ represents the tokenized answer sequence provided by either ECG-QA \citep{oh2023ecgqacomprehensivequestionanswering} or MIMIC-IV ECG pretraining \citep{zhao2024ecgchatlargeecglanguagemodel}.

Given these two features $I$ and $O$ we then describe the contrastive, masked, and dual approaches implemented for our baselines that are derived from commonly used techniques used throughout previous works \citep{oh2022leadagnosticselfsupervisedlearninglocal, choi2023ecgbert, mckeen2024ecgfmopenelectrocardiogramfoundation, pham2024cmeltcontrastiveenhancedmasked, tang2024electrocardiogramlanguagemodelfewshotquestion, tang2024electrocardiogramreportgenerationquestion, vaid2022heartbeit}.

\subsection{Contrastive learning approaches}\label{apd:cont-learn}
We utilize a pretrained CLIP \cite{Radford2021LearningTV} checkpoint, namely `openai/clip-vit-base-patch32', provided by HuggingFace \citep{wolf2020huggingfaces} to encode ECG signals $I$ and text labels $O$ into a shared embedding space. 
Let $f_{\text{img}} : \mathbb{R}^{3 \times C \times T} \to \mathbb{R}^d$ and $f_{\text{txt}} : \text{Text} \to \mathbb{R}^d$ be the image and text encoders of the pretrained CLIP model, respectively. The embeddings for the $i$-th pair are computed as:
\[
z_i^{\text{img}} = f_{\text{img}}(I_i), \quad z_i^{\text{txt}} = f_{\text{txt}}(O_i),
\]
where $z_i^{\text{img}}, z_i^{\text{txt}} \in \mathbb{R}^d$.
The CLIP loss function \(\mathcal{L}_{\text{CLIP}}\) aligns the embeddings of corresponding ECG signals and text labels while contrasting them with non-matching pairs. This is formulated as:
\begin{equation}
\resizebox{\columnwidth}{!}{$
\mathcal{L}_{\text{CL}} = -\frac{1}{N} \sum_{i=1}^N \Biggl[
    \log \frac{\exp\left(\text{sim}(z_i^{\text{img}}, z_i^{\text{txt}})/\tau\right)}
         {\sum_{j=1}^N \exp\left(\text{sim}(z_i^{\text{img}}, z_j^{\text{txt}})/\tau\right)}
    + \log \frac{\exp\left(\text{sim}(z_i^{\text{txt}}, z_i^{\text{img}})/\tau\right)}
         {\sum_{j=1}^N \exp\left(\text{sim}(z_i^{\text{txt}}, z_j^{\text{img}})/\tau\right)}
\Biggr]
$}
\end{equation}
where \(\text{sim}(\cdot, \cdot)\) denotes cosine similarity, and \(\tau\) is a learnable temperature parameter.

To integrate the pretrained CLIP model into our language model for joint reasoning over ECG signals and text, we project the frozen image embeddings $z_i^{\text{img}}$ into the language model's hidden space. Let $W \in \mathbb{R}^{h \times d}$ be a learnable projection matrix, where $h$ is the hidden dimension of the language model. The projected embeddings are:
\[
z_i^{\text{clip}} = W z_i^{\text{img}}.
\]
These projected embeddings $z_i^{\text{clip}}$ are then prepended to the token embeddings of the language model, where we get \( \texttt{Context} = \{ \texttt{[BOS]}, \texttt{[SIG\_START]}, z_i^{\text{clip}}, \texttt{[SIG\_END]}, Q \} \) to train the same autoregressive objective, $L_{NLL}$.

\subsection{Masked image modeling approaches}\label{apd:mlm-learn}
Consider the normalized ECG image $I \in \mathbb{R}^{3 \times C \times T}$ obtained as previously described. We utilize a pretrained Vision Transformer (ViT) model \citep{dosovitskiy2021image}, specifically the `google/vit-base-patch16-224-in21k' checkpoint provided by HuggingFace \citep{wolf2020huggingfaces}.

The image $I$ is partitioned into $P$ non-overlapping patches. Let $N$ be the number of images in our dataset, and $I_i$ denote the $i$-th image. The ViT encoder $f_{\text{vit}}$ projects these patches into latent embeddings:
\[
z_i^{\text{patch}} = f_{\text{vit}}(I_i) \in \mathbb{R}^{P \times d},
\]
where $d$ is the embedding dimension of the ViT model.

During training, we randomly mask a subset of patches for each image $I_i$, creating a binary mask $M_i \in \{0,1\}^P$, where $M_{i,j} = 1$ if patch $j$ is masked and $M_{i,j} = 0$ otherwise. The masked embeddings $z_i^{\text{masked}}$ are formed by replacing the embeddings of masked patches with a mask token. A reconstruction head $f_{\text{rec}}$ is then applied to predict the pixel-level content of the masked patches:
\[
\hat{I}_i = f_{\text{rec}}(z_i^{\text{masked}}) \in \mathbb{R}^{P \times d}.
\]
The masked image modeling loss $\mathcal{L}_{\text{MIM}}$ is computed as the mean squared error (MSE) between the reconstructed embeddings $\hat{I}_i$ and the original embeddings $z_i^{\text{patch}}$ at the masked positions:
\begin{equation}
\mathcal{L}_{\text{MIM}} = \frac{1}{N} \sum_{i=1}^N \frac{1}{\sum_{j=1}^P M_{i,j}} \sum_{j=1}^P M_{i,j} \left\| \hat{I}_i[j] - z_i^{\text{patch}}[j] \right\|_2^2.
\label{eq:mim_loss}
\end{equation}
To integrate the MIM representations into the language model for joint reasoning over ECG signals and textual questions, we project the frozen ViT embeddings $z_i^{\text{img}} \in \mathbb{R}^d$ into the language model's hidden space. Let $W \in \mathbb{R}^{h \times d}$ be a learnable projection matrix, where $h$ is the hidden dimension of the language model. The projected embeddings are given by:
\[
z_i^{\text{vit}} = W z_i^{\text{img}}.
\]
These projected embeddings $z_i^{\text{vit}}$ are then prepended to the language model's token embeddings, to get \( \texttt{Context} = \{ \texttt{[BOS]}, \texttt{[SIG\_START]}, z_i^{\text{vit}}, \texttt{[SIG\_END]}, Q \} \) to train the same autoregressive objective, $L_{NLL}$, mentioned previously. 

\begin{algorithm}[t]
\caption{Merging a Pair in an ID Array \textit{merge}}
{\bfseries Input:} Array of IDs \texttt{ids}, pair to merge \texttt{pair} as \((u_1, u_2)\), new ID \texttt{new\_id}. \par
{\bfseries Output:} Updated array of IDs \texttt{ids} with merged pairs. \par
\begin{algorithmic}[1]
    \STATE \(i \gets 0\), \(write \gets 0\)
    \WHILE{\(i < \text{len(\texttt{ids})}\)}
        \IF{\(i + 1 < \text{len(\texttt{ids})}\) \textbf{and} \((\texttt{ids}[i], \texttt{ids}[i+1]) = \texttt{pair}\)}
            \STATE Set \(\texttt{ids}[write] \gets \texttt{new\_id}\)
            \STATE \(write \gets write + 1\)
            \STATE \(i \gets i + 2\)
        \ELSE
            \STATE Set \(\texttt{ids}[write] \gets \texttt{ids}[i]\)
            \STATE \(write \gets write + 1\)
            \STATE \(i \gets i + 1\)
        \ENDIF
    \ENDWHILE
    \STATE Truncate \texttt{ids} to length \(write\)
    \STATE \textbf{return} \texttt{ids}
\end{algorithmic}
\label{alg:merge}
\end{algorithm}

\subsection{Dual approaches}\label{apd:cont-mlm-learn}
The dual approach follows the previous two contrastive and masked image modeling approaches for pretraining the ECG encoder but simply just combines the losses like so:
\[
\mathcal{L}_{\text{Dual}} = \lambda_1\mathcal{L}_{\text{MIM}} + \lambda_2\mathcal{L}_{\text{CL}}
\]
where $\lambda_1 = \lambda_2 = 1$ in our study.

However, when training the autoregressive LLM, we project both embeddings, $z_i^{\text{vit}}$ and $z_i^{\text{clip}}$, outputted by their respective frozen encoders via a learnable projection matrix into the language model's hidden space of dimension $h$.
We then concatenate the projected embeddings and pass them through a fusion network to obtain the fused visual embedding $z_i^{\text{fused}} \in \mathbb{R}^h$:
\[
z_i^{\text{fused}} = f_{\text{fusion}}(\text{concat}(z_i^{\text{vit}}; z_i^{\text{clip}})),
\]
where $f_{\text{fusion}}$ is a trainable feedforward network.
The fused visual embedding $z_i^{\text{fused}}$ is prepended to the token embeddings of the language model, forming \( \texttt{Context} = \{ \texttt{[BOS]},\\ \texttt{[SIG\_START]}, z_i^{\text{fused}}, \texttt{[SIG\_END]}, Q \} \) to train the autoregressive objective, $L_{NLL}$.

\begin{algorithm}[t]
\caption{Calculating Frequency of Byte Pairs in an Array \textit{get\_stats}}
{\bfseries Input:} Array of IDs \texttt{ids}. \par
{\bfseries Output:} HashMap of byte pairs and their frequencies \texttt{pair\_counts}. \par
\begin{algorithmic}[1]
    \IF{\(\text{len(ids)} < 1000\)}
        \STATE \texttt{pair\_counts} \(\gets\) Empty HashMap
        \FORALL{each window of size 2 in \texttt{ids}}
            \STATE \((u_1, u_2) \gets\) two elements of the window
            \STATE Increment count for \((u_1, u_2)\) in \texttt{pair\_counts}
        \ENDFOR
    \ELSE
        \STATE \texttt{pair\_counts} \(\gets\) Parallel \textit{fold} operation:
        \FORALL{each window of size 2 in \texttt{ids} (in parallel)}
            \STATE \((u_1, u_2) \gets\) two elements of the window
            \STATE Increment count for \((u_1, u_2)\) in the local HashMap
        \ENDFOR
        \STATE Combine local HashMaps using a parallel \textit{reduce} operation to obtain \texttt{pair\_counts}
    \ENDIF
    \STATE \textbf{return} \texttt{pair\_counts}
\end{algorithmic}
\label{alg:getstats}
\end{algorithm}

\section{Additional Results and Discussions}\label{apd:third}

\begin{table*}[hbtp]
  \caption{Mean results with standard deviations over 5 random seeds on zero shot cross-dataset transferability.}
  {\resizebox{\textwidth}{!}{%
  \small \begin{tabular}{lllcccc}
  \toprule
  \bfseries Method & \bfseries Trained Dataset & \bfseries Inferenced Dataset& \bfseries BLEU-4 & \bfseries Rouge-L & \bfseries Meteor & \bfseries BertScore F1 \\
  \midrule
  $L_{CL}$ & \multirow{5}{*}{ECG-QA MIMIC-IV} & \multirow{5}{*}{ECG-QA PTB-XL} & 11.64 ± 0.45 & 41.48 ± 0.11 & 25.74 ± 0.13 & 91.24 ± 0.05 \\
  $L_{MIM}$ &  & & \textbf{11.70} ± 0.29 & \textbf{42.22} ± 0.28 & \textbf{26.41} ± 0.10 & 91.51 ± 0.03 \\
  $L_{MERL}$ \citep{liu2024zeroshotecgclassificationmultimodal}& & & 11.53 ± 0.19 & 39.23 ± 0.40 & 25.58 ± 0.28 & \textbf{91.59} ± 0.03 \\
  $L_{Dual}$ &  & & 9.71 ± 0.10 & 35.10 ± 0.28 & 24.91 ± 0.19  & 87.88 ± 0.08 \\
  \textbf{ECG-Byte} &  & & 8.70 ± 0.04 & 40.39 ± 0.40 & 23.29 ± 0.18 & 91.51 ± 0.03 \\
  \midrule
  $L_{CL}$ & \multirow{5}{*}{ECG-QA PTB-XL} & \multirow{5}{*}{ECG-QA MIMIC-IV} & 5.10 ± 0.04 & 22.77 ± 0.28 & 14.63 ± 0.32 & 77.89 ± 0.13 \\
  $L_{MIM}$ &  & & 7.68 ± 0.46 & \textbf{35.77} ± 0.13 & \textbf{22.32} ± 0.33 & 90.28 ± 0.07 \\
  $L_{MERL}$ \citep{liu2024zeroshotecgclassificationmultimodal}&  & & 7.39 ± 0.15 & 28.33 ± 0.58 & 18.59 ± 0.35 & 89.30 ± 0.05 \\
  $L_{Dual}$ &  & & 7.49 ± 0.21 & 30.53 ± 0.59 & 20.25 ± 0.27 & 86.53 ± 0.11 \\
  \textbf{ECG-Byte} &  & & \textbf{7.86} ± 0.13 & 35.01 ± 0.41& 21.49 ± 0.24& \textbf{90.29} ± 0.07\\
  \bottomrule
  \end{tabular}}}
  \label{tab:cross}
\end{table*}

\subsection{Cross Dataset Transferability}
We present the results of cross-dataset transferability in Table~\ref{tab:cross}, comparing our approach, \textbf{ECG-Byte}, with two-stage pretraining methods. 
\textbf{ECG-Byte} achieves the best zero-shot transfer performance in BLEU-4 and BertScore F1 scores when transferring from the ECG-QA PTB-XL dataset to the ECG-QA MIMIC-IV dataset. 
When transferring from the ECG-QA MIMIC-IV dataset to the ECG-QA PTB-XL dataset, although other two-stage pretraining methods demonstrate higher performance, \textbf{ECG-Byte} maintains competitive results across all metrics.

\begin{table}[!hbtp]
\centering
  \caption{Ablation study on varying number of merges $\texttt{num\_merges}$.}
  {\resizebox{0.8\columnwidth}{!}{%
  \small \begin{tabular}{lcccc}
  \toprule
  \bfseries $\texttt{num\_merges}$ & \bfseries BLEU-4 & \bfseries Rouge-L & \bfseries Meteor & \bfseries BertScore F1 \\
  \midrule
  500 & 13.61 ± 0.53 & 46.50 ± 0.28 & 28.49 ± 0.49 & 92.33 ± 0.02 \\
  1750 & 14.50 ± 0.25 & 46.74 ± 0.48 & 30.03 ± 0.25 & \textbf{92.55} ± 0.01 \\
  2500 & \textbf{15.10} ± 0.39 & 46.37 ± 0.28 & \textbf{30.12} ± 0.23  & 92.53 ± 0.05 \\
  3500 & 13.93 ± 0.21 & \textbf{47.08} ± 0.56 & 29.17 ± 0.31 & 92.53 ± 0.07 \\
  \bottomrule
  \end{tabular}}}
  \label{tab:num_merges}
\end{table}

\paragraph{Number of Merges}
The number of merges \texttt{num\_merges} performed during training \textbf{ECG-Byte} corresponds to how much the algorithm compresses the concatenated sequence of quantized ECGS $\mathbf{X}_{\text{concat}}$.
More \texttt{num\_merges} means more compression, which can affect the expressiveness of the encoded sequence.
In Table~\ref{tab:num_merges}, we show the performance of our method with different \texttt{num\_merges}.
The results suggest that, although performance fluctuates slightly with the number of merges, varying \texttt{num\_merges} generally produces comparable outcomes.

\subsection{Does Larger LLMs Yield Higher Performance?}
We present the results of ablating the size of the LLM in Table~\ref{tab:large_llm}. 
Interestingly, the performance across the three different model sizes (1B, 3B, 8B) remains fairly similar. 
We believe that the limited dataset size prevents the larger models from realizing their full performance potential. 
We hypothesize that increasing the amount of training data would enable the larger models to leverage their greater capacity.

\begin{table}[!hbtp]
  \caption{Ablation study on how larger LLMs perform for NLG.}
  {\resizebox{\columnwidth}{!}{%
  \small \begin{tabular}{lcccc}
  \toprule
  \bfseries LLM & \bfseries BLEU-4 & \bfseries Rouge-L & \bfseries Meteor & \bfseries BertScore F1 \\
  \midrule
  Llama 3.2 1B \citep{grattafiori2024llama3herdmodels}&13.93 ± 0.21 & \textbf{47.08} ± 0.56 & 29.17 ± 0.31 & \textbf{92.53} ± 0.07 \\
  Llama 3.2 3B \citep{grattafiori2024llama3herdmodels}& \textbf{14.80} ± 0.17 & 46.55 ± 0.21 & \textbf{29.53} ± 0.16 & 92.42 ± 0.01 \\
  Llama 3.1 8B \citep{grattafiori2024llama3herdmodels}& 13.80 ± 0.16 & 46.29 ± 0.25 & 28.56 ± 0.11 & 92.44 ± 0.05 \\
  \bottomrule
  \end{tabular}}}
  \label{tab:large_llm}
\end{table}

\subsection{Encoder-Free Vision-Language Models}
Released in 2023, Fuyu-8B \citep{bavishi_2023_fuyu8b} was a seminal work that introduced the concept of encoder-free Vision-Language Models (VLMs). To align the image modality with text, Fuyu-8B first patches an image and then projects these patches linearly into the first layer of a decoder-only transformer network \citep{vaswani2023attention}. Fuyu-8B demonstrated strong performance on visual understanding tasks, and since its release, several works have explored encoder-free architectures for VLMs \citep{diao2024unveilingencoderfreevisionlanguagemodels, wang2025visionlora}. However, at the time of this study, most VLMs continue to rely on vision encoders, so we did not pursue additional experiments applying encoder-free methods to ECGs.
Nonetheless, there is a clear parallel in motivation between encoder-free VLMs and \textbf{ECG-Byte}. Utilizing ECG-specific encoders—whether explicitly designed for ECGs or pretrained on internet-scale data—can introduce strong inductive biases into the learned representations \citep{diao2024unveilingencoderfreevisionlanguagemodels}. In the encoder-free VLM literature, simple and low-overhead neural networks such as linear projections \citep{bavishi_2023_fuyu8b}, MLPs \citep{wang2025visionlora}, and convolutional networks \citep{diao2024unveilingencoderfreevisionlanguagemodels} are used to align visual inputs with language models.
In contrast, \textbf{ECG-Byte} employs a purely rule-based method—Byte Pair Encoding (BPE) \citep{Gage1994ANA}—to tokenize ECGs directly, enabling seamless integration with language model input alongside text tokens. While learning-based compressors with low overhead as seen in the encoder-free VLM literature may hold potential, we observe that even without any learnable modules to encode the ECG signal, \textbf{ECG-Byte} achieves strong performance. Therefore, we leave the exploration of incorporating encoder-free VLM concepts into the development of Electrocardiogram-Language Models (ELMs) to future work.

\subsection{Qualitative NLG Examples}\label{apd:nlg}
\begin{figure*}[htp]
\centering
\includegraphics[width=1\linewidth]{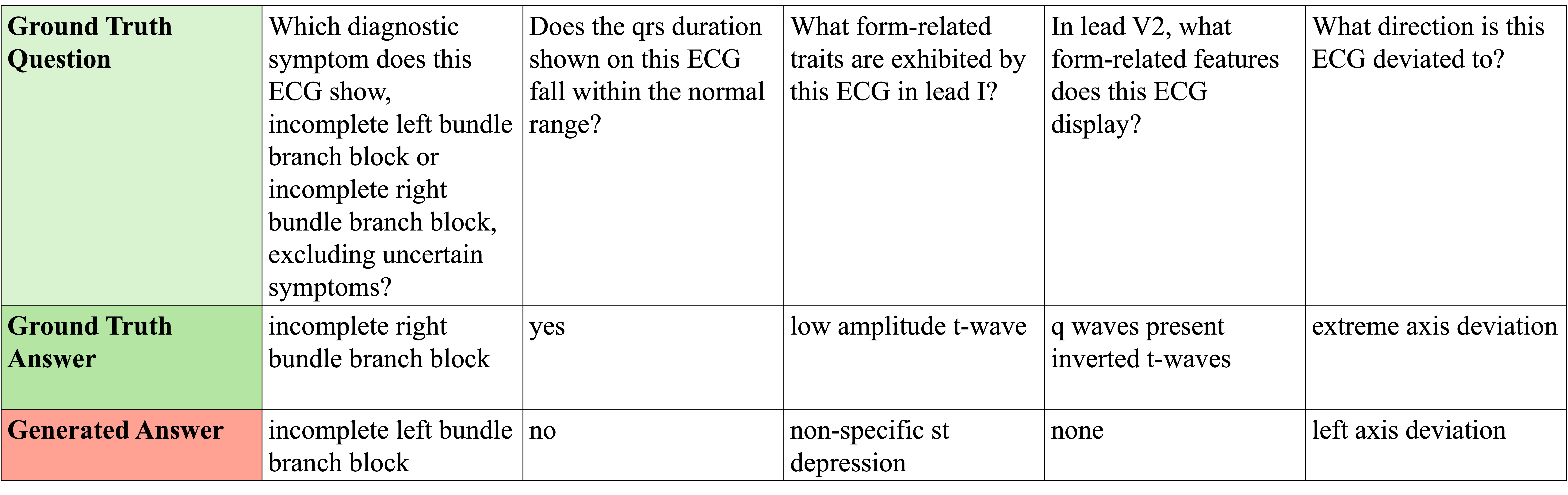}
\includegraphics[width=1\linewidth]{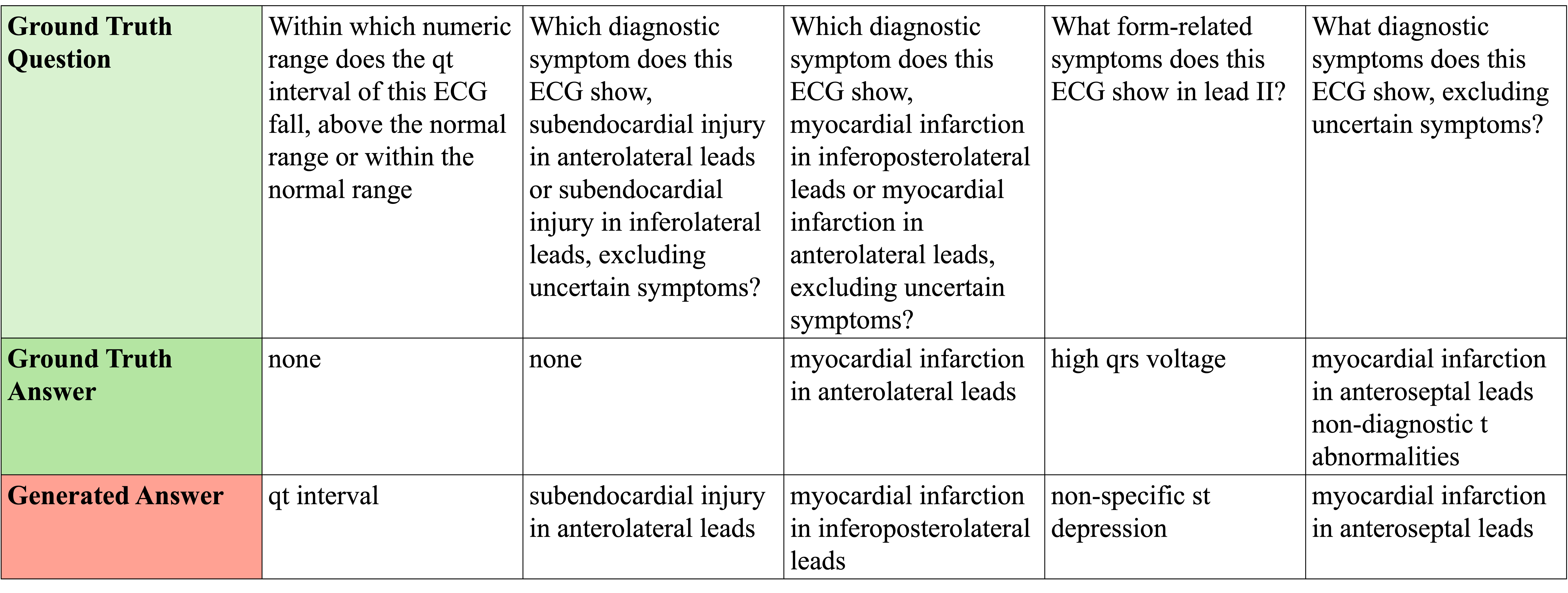}
\caption{Randomly sampled NLG results of unsuccessful generations on the PTB-XL test set from ECG-QA.}
\label{Fig:results_gen2}
\end{figure*}

\begin{figure*}[htp]
\centering
\includegraphics[width=1\linewidth]{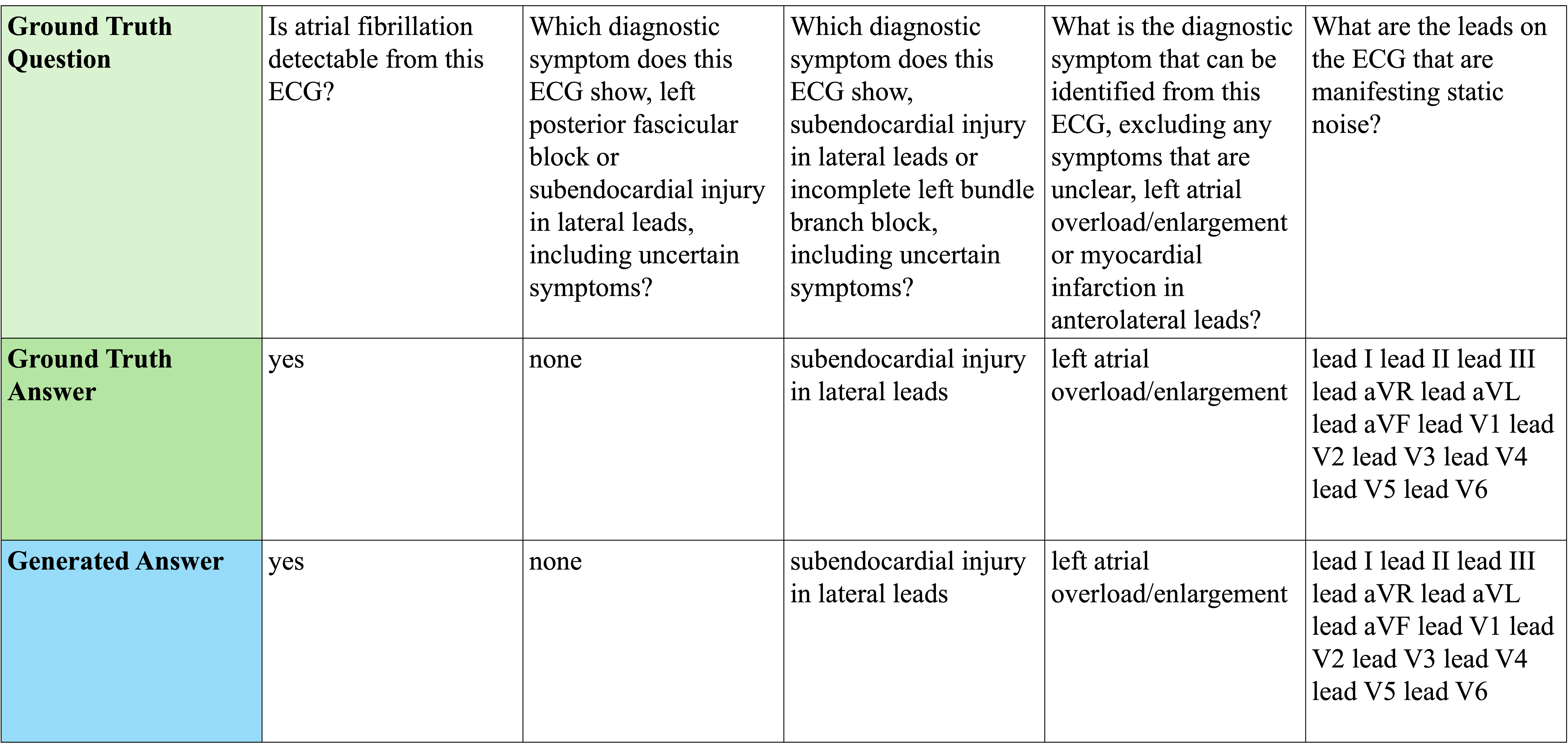}
\includegraphics[width=1\linewidth]{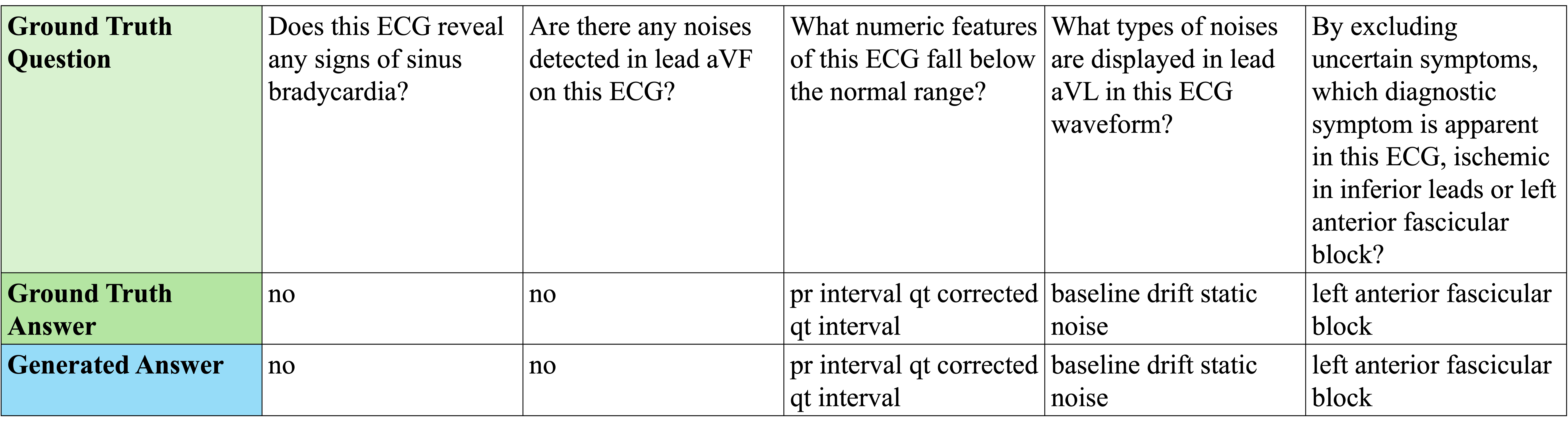}
\caption{Randomly sampled NLG results of successful generations on the PTB-XL test set from ECG-QA.}
\label{Fig:results_gen}
\end{figure*}

We provide qualitative NLG examples of successful (Figure~\ref{Fig:results_gen}) and unsuccessful generations (Figure~\ref{Fig:results_gen2}).

\clearpage
\begin{algorithm}[t]
\caption{Encoding Process for \textbf{ECG-Byte}}
{\bfseries Input:} Input \(X_{\text{symb}}\) and merge history \texttt{merges}. \par
{\bfseries Output:} Vector of encoded IDs \texttt{output\_ids}. \par
\begin{algorithmic}[1]
    \STATE \texttt{ids} \(\gets\) Convert \(X_{\text{symb}}\) to a vector of \texttt{u32}
    \STATE \texttt{trie\_root} $\gets$ Initialize root TrieNode
    \FOR{\(b \gets 0\) \textbf{to} \(255\)}
        \STATE \textit{insert}(\texttt{trie\_root}, [\texttt{b}], \texttt{b})
    \ENDFOR
    \FOR{\textbf{each} \((\texttt{token\_sequence}, \texttt{token\_id})\) in \texttt{merges}}
        \STATE \textit{insert}(\texttt{trie\_root}, \texttt{token\_sequence}, \\ \texttt{token\_id})
    \ENDFOR
    \STATE \texttt{output\_ids} $\gets$ Empty list
    \STATE \(i \gets 0\)
    \WHILE{\(i < \text{len(\texttt{ids})}\)}
        \STATE \(\texttt{node} \gets \texttt{trie\_root}\)
        \STATE \(\texttt{match\_len} \gets 0\)
        \STATE \(\texttt{match\_id} \gets \text{None}\)
        \FOR{\(j \gets i\) \textbf{to} \(\text{len(\texttt{ids})} - 1\)}
            \STATE \(\texttt{id} \gets \texttt{ids}[j]\)
            \IF{\(\texttt{id}\) exists in \(\texttt{node.children}\)}
                \STATE \(\texttt{node} \gets \texttt{node.children[id]}\)
                \IF{\(\texttt{node.token\_id} \neq \text{None}\)}
                    \STATE \(\texttt{match\_len} \gets j - i + 1\)
                    \STATE \(\texttt{match\_id} \gets \texttt{node.token\_id}\)
                \ENDIF
            \ELSE
                \STATE \textbf{break}
            \ENDIF
        \ENDFOR
        \IF{\(\texttt{match\_id}\neq \texttt{None}\)}
            \STATE Append \(\texttt{match\_id}\) to \texttt{output\_ids}
            \STATE \(i \gets i + \texttt{match\_len}\)
        \ELSE
            \STATE Append \(\texttt{ids}[i]\) to \texttt{output\_ids}
            \STATE \(i \gets i + 1\)
        \ENDIF
    \ENDWHILE
    \STATE \textbf{return} \texttt{output\_ids}
\end{algorithmic}
\label{algo:ecg_byte_encode}
\end{algorithm}
\clearpage
\clearpage
\clearpage
\clearpage

\end{document}